\documentclass[letterpaper]{article} 
\usepackage{aaai24}  
\usepackage{xcolor}
\usepackage{times}  
\usepackage{helvet}  
\usepackage{courier}  
\usepackage[hyphens]{url}  
\usepackage{graphicx} 
\usepackage{natbib}  
\usepackage{caption} 
\usepackage{algorithm}
\usepackage{algorithmic}

\usepackage{booktabs} 
\usepackage{subcaption}

\usepackage{newfloat}
\usepackage{listings}
\DeclareCaptionStyle{ruled}{labelfont=normalfont,labelsep=colon,strut=off} 
\floatstyle{ruled}
\newfloat{listing}{tb}{lst}{}
\floatname{listing}{Listing}
\def  \shortname {\textit{TactileTango }}
\def  \framework {\textit{Prompt-Propose-Verify }}
\usepackage{array}  
\usepackage{lipsum}  

\title{\textit{Prompt-Propose-Verify}: A Reliable Hand-Object-Interaction Data Generation Framework using Foundational Models}
\author {
    Gurusha Juneja\textsuperscript{\rm 1}\equalcontrib,
    Sukrit Kumar\textsuperscript{\rm 2}\equalcontrib
}
\affiliations {
    \textsuperscript{\rm 1}IIT Delhi\\
    \textsuperscript{\rm 2}BITS Pilani, Hyderabad\\
    ee1190480@iitd.ac.in, ksukrit2001@gmail.com
}

\usepackage{bibentry}

\begin{document}

\maketitle

\begin{abstract}
Diffusion models when conditioned on text prompts, generate realistic-looking images with intricate details. But most of these pre-trained models fail to generate accurate images when it comes to human features like hands, teeth, etc. We hypothesize that this inability of diffusion models can
be overcome through well-annotated good-quality data. In this paper, we look specifically into improving the hand-object-interaction image generation using diffusion models. We collect a well annotated hand-object interaction synthetic dataset curated using \framework framework and finetune a stable diffusion model on it. We evaluate
the image-text dataset on qualitative and quantitative metrics like CLIPScore, ImageReward, Fedility, and alignment and show considerably better performance over the current state-of-the-art benchmarks.

\end{abstract}

\section{Introduction}

\begin{table*}[hbtp]
    \centering
    \begin{tabular}{|m{3cm}|c|c|c|}
    \hline
         \textbf{Prompt} & \textbf{Stable Diffusion V1.4} & \textbf{DALL·E 2} & \textbf{StableDiffusionXL}\\
    \hline
         \begin{minipage}{\linewidth}
 A poster of a car race with the words "Monaco" on it.
        \end{minipage}
        &
        \begin{minipage}{0.2\linewidth} \includegraphics[width=\textwidth]{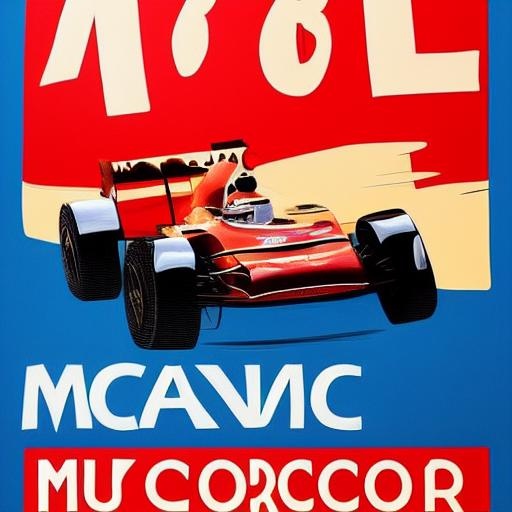} \end{minipage} & 
        \begin{minipage}{0.2\linewidth} \includegraphics[width=\textwidth]{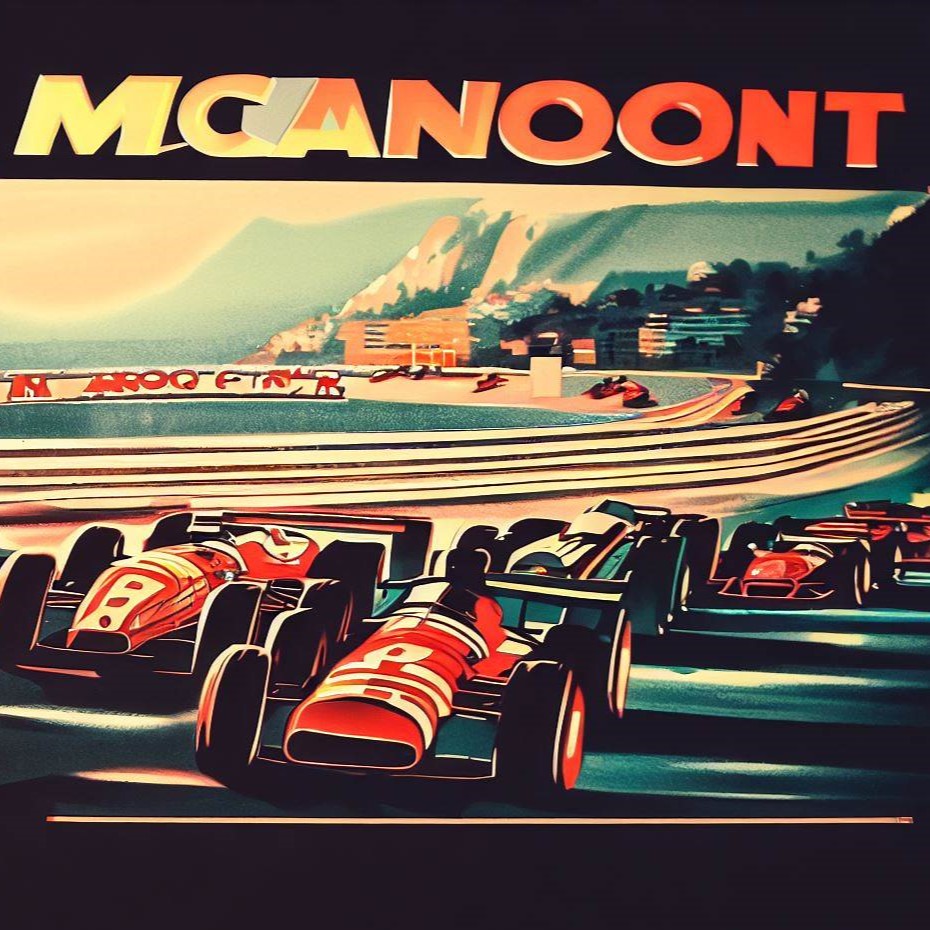} \end{minipage}&  
        \begin{minipage}{0.2\linewidth} \includegraphics[width=\textwidth]{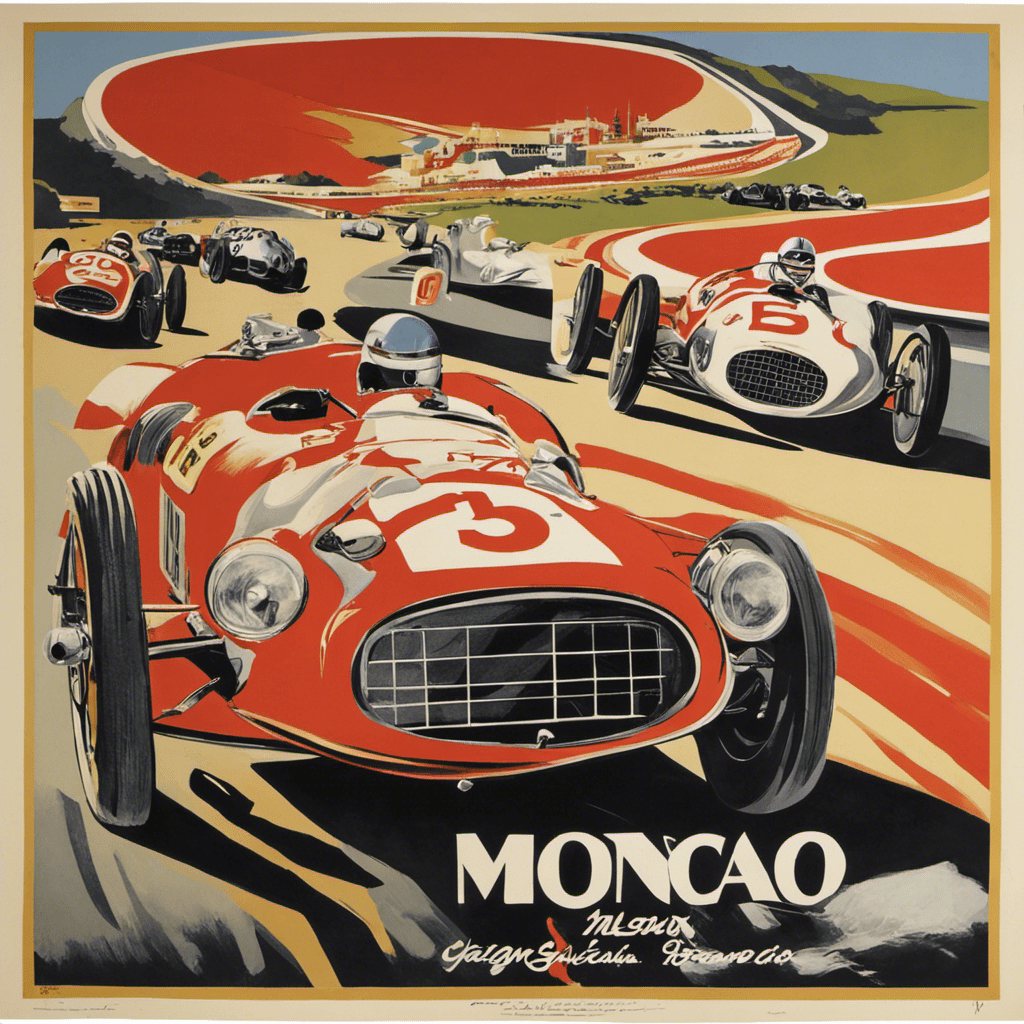} \end{minipage}\\

  \hline      
        \begin{minipage}{\linewidth}
A human hand with a ring on the ring finger resting on a plain surface
        \end{minipage}
        &
        \begin{minipage}{0.2\linewidth} \includegraphics[width=\textwidth]{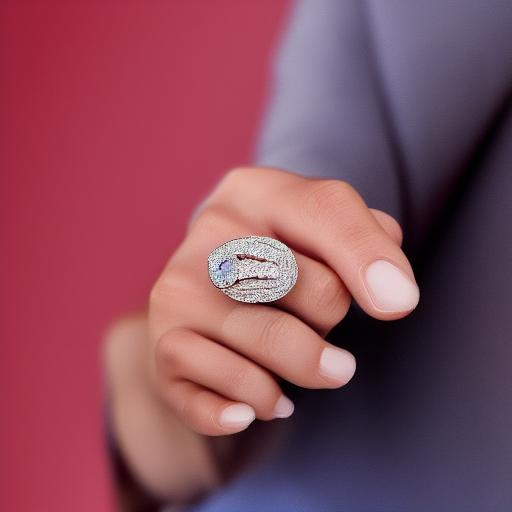} \end{minipage} & 
        \begin{minipage}{0.2\linewidth} \includegraphics[width=\textwidth]{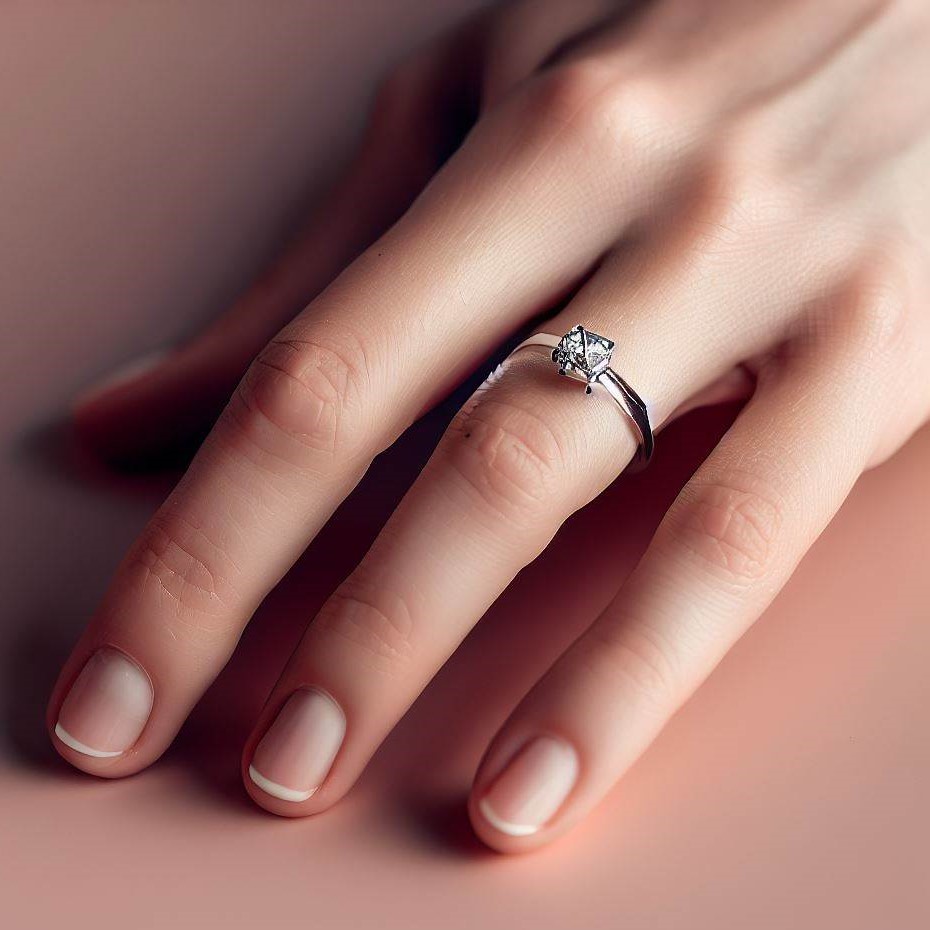} \end{minipage}&  
        \begin{minipage}{0.2\linewidth} \includegraphics[width=\textwidth]{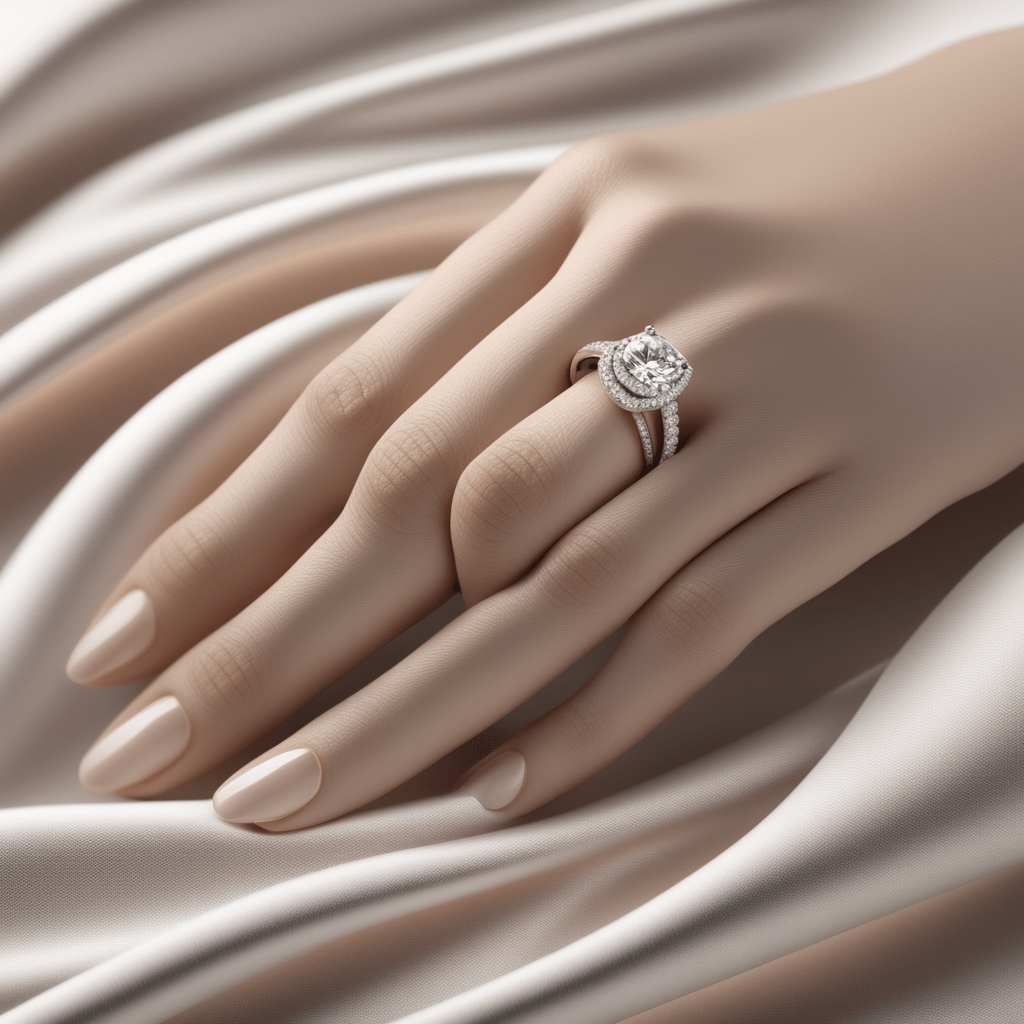} \end{minipage}\\
   
  \hline
        
    \end{tabular}
    \caption{Sample prompts where diffusion models fail to generate good realistic looking-images.}
    \label{tab:failre_modes}
\end{table*}

 \begin{figure*}[hbtp]
    \centering
    \begin{subfigure}[t]{0.3\linewidth}
        \includegraphics[width=\linewidth]{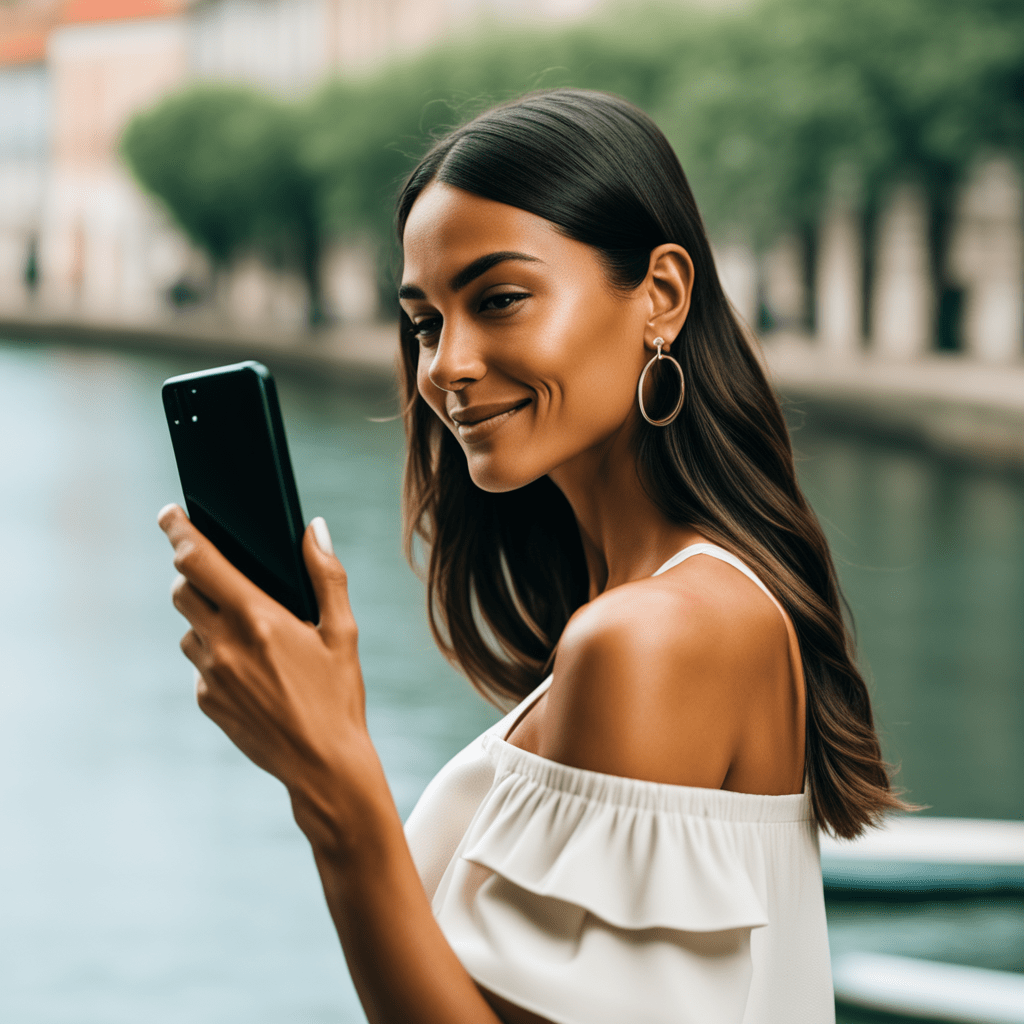}
        \caption{A realistic image of a woman with slightly tanned hands. Left Hand depicts a finger grasp motion. her left hand securely holding a phone, thumb fully open resting on the edge and fingers gently half curled around the back. Palm not touching phone.}
    \end{subfigure}
    \begin{subfigure}[t]{0.3\linewidth}
        \includegraphics[width=\linewidth]{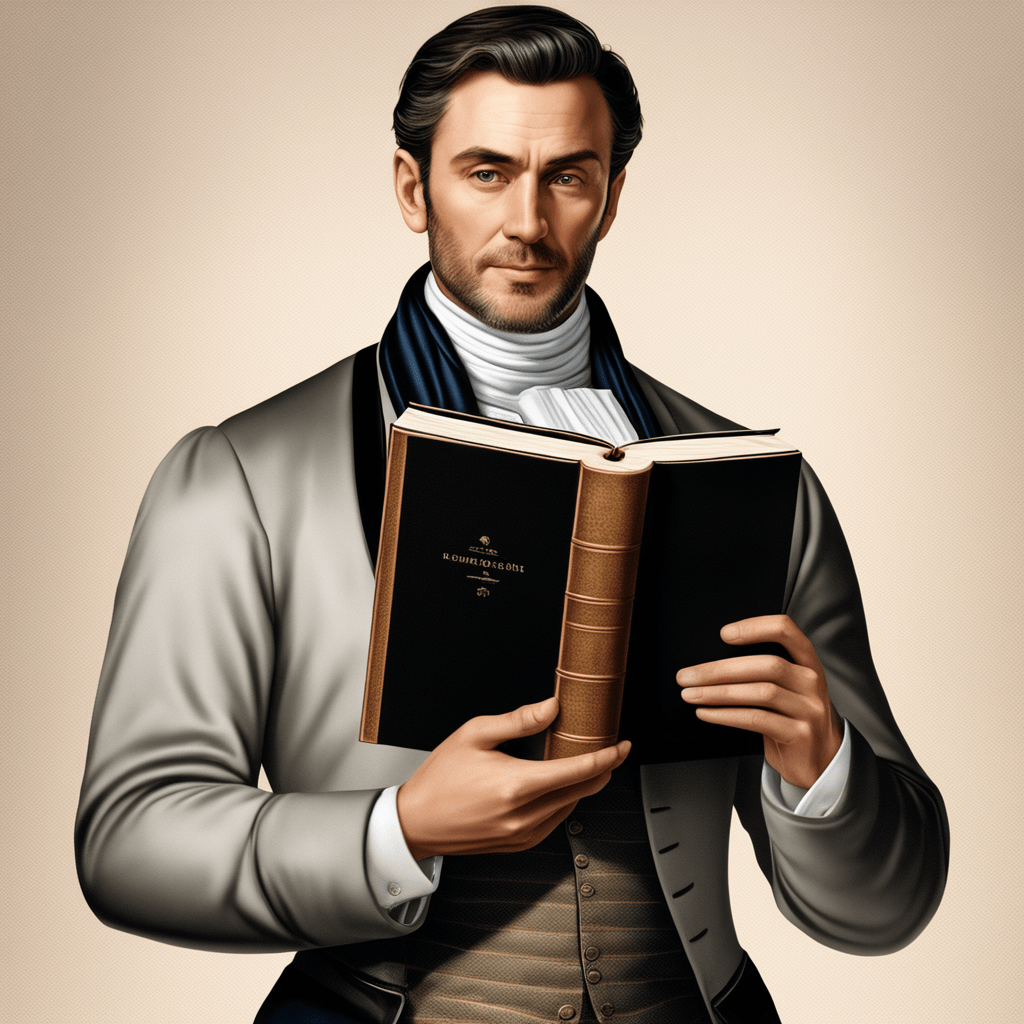}
        \caption{A realistic image of a man's right hand holding a book, with the thumb fully open on the bottom edge, index finger fully open on the spine, and other fingers also fully open spread across the back cover. The book appears slightly open, revealing pages.}
    \end{subfigure}
    \begin{subfigure}[t]{0.3\linewidth}
        \includegraphics[width=\linewidth]{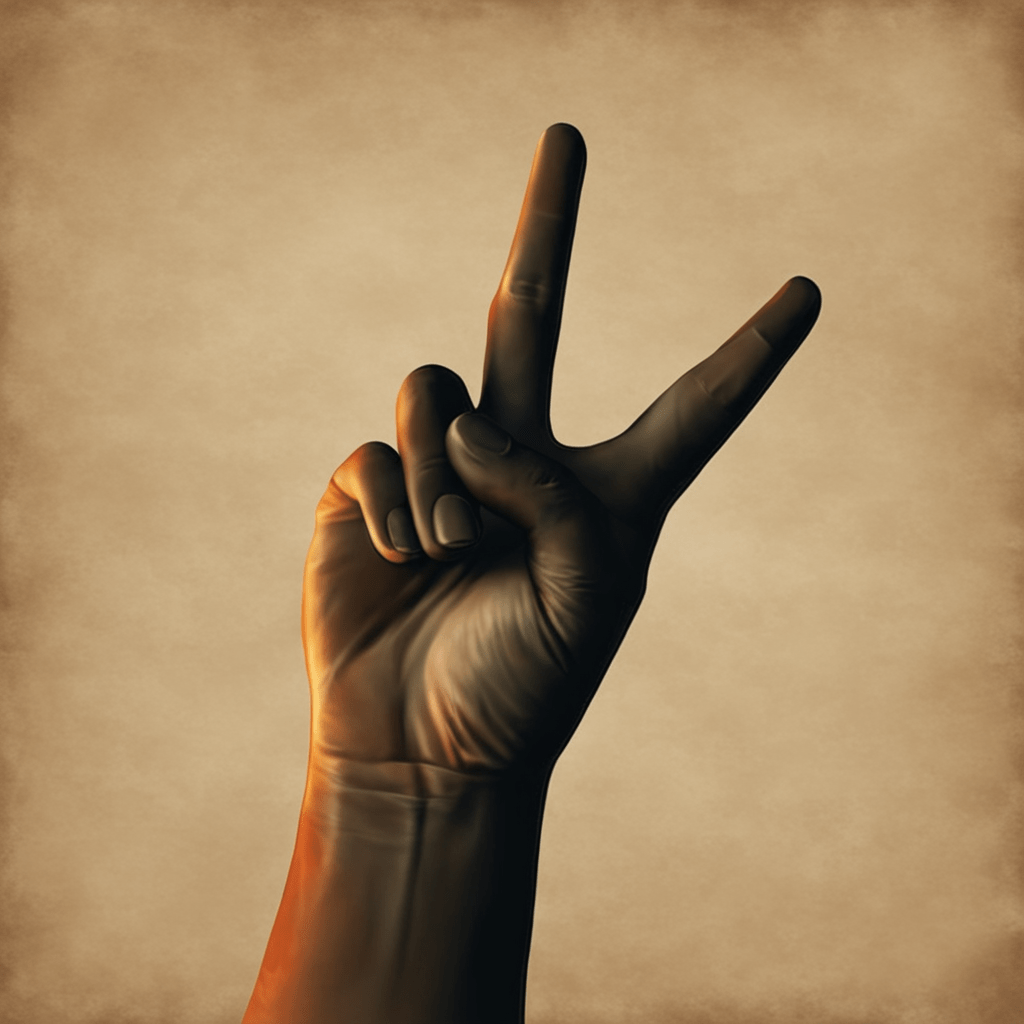}
        \caption{A realistic, high-resolution image of a human hand displaying the peace sign, with the index and middle fingers fully open extended upward, and the thumb, ring, and pinky fingers fully curled inward, with clear finger positioning and hand gesture}
    \end{subfigure}
    
    \begin{subfigure}[t]{0.3\linewidth}
        \includegraphics[width=\linewidth]{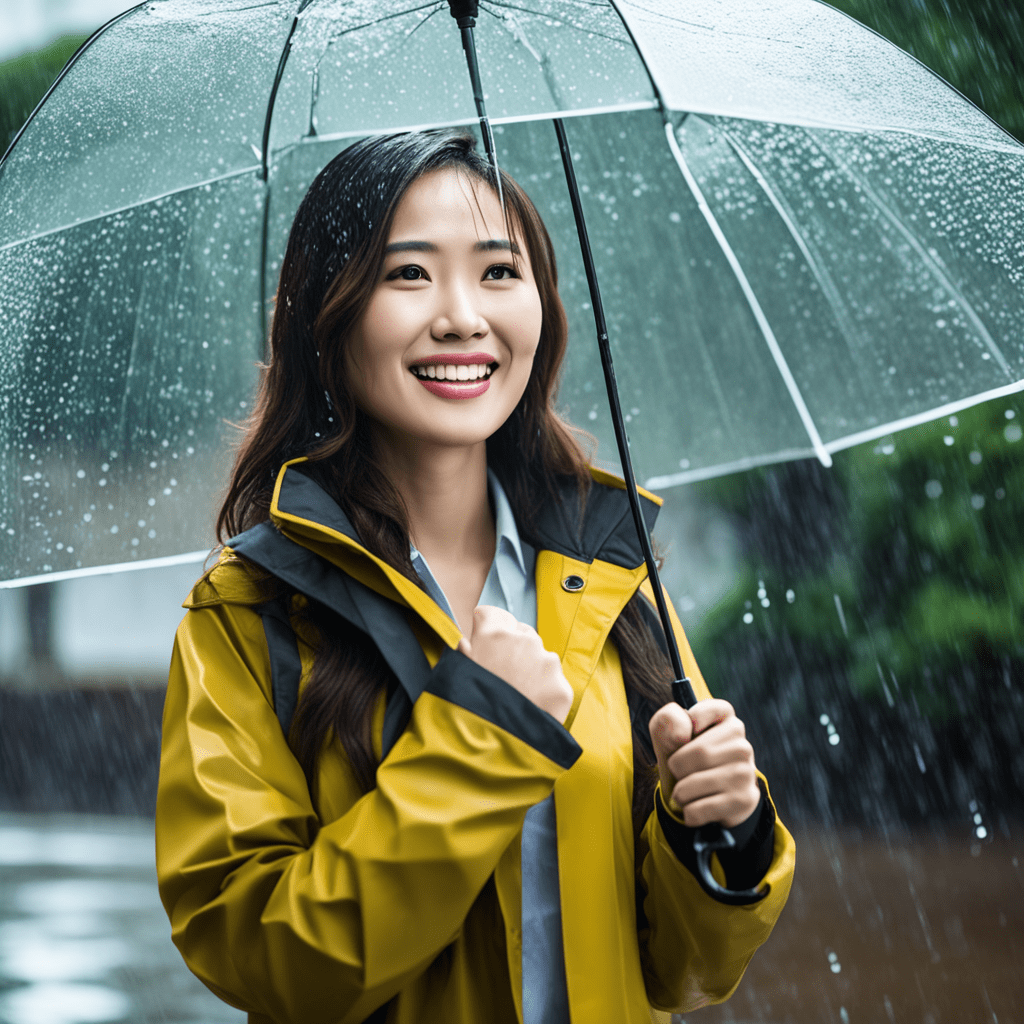}
        \caption{A high-resolution image of a female's left hand in a full hand grasp motion gracefully holding an umbrella handle, with her thumb and fingers encircling it in a fully closed position. Raindrops gently fall on the umbrella, creating a serene atmosphere}
    \end{subfigure}
    \begin{subfigure}[t]{0.3\linewidth}
        \includegraphics[width=\linewidth]{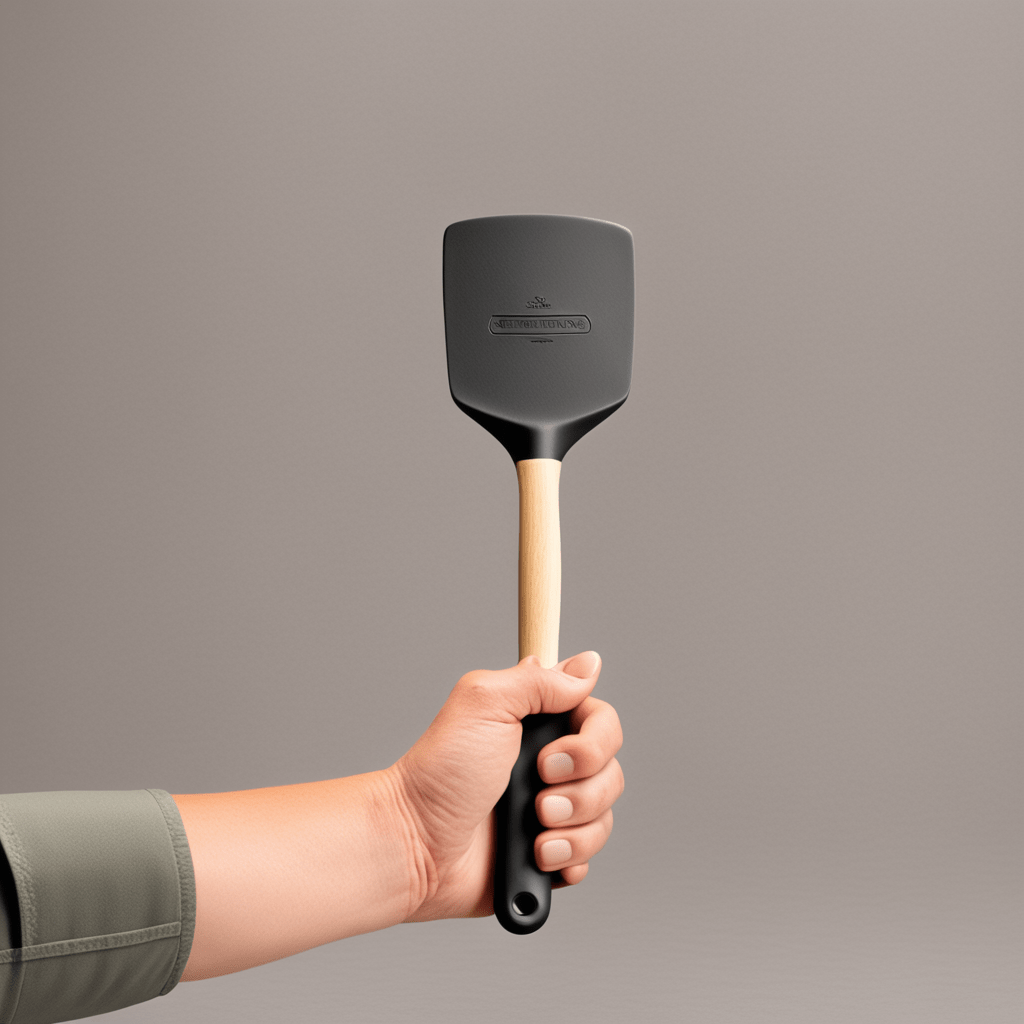}
        \caption{A realistic human hand holds a kitchen spatula, with the thumb and index finger fully curled around the handle while the middle, ring, and pinky fingers rest along its length, half curled. The spatula's flat surface faces upward}
    \end{subfigure}
    \begin{subfigure}[t]{0.3\linewidth}
        \includegraphics[width=\linewidth]{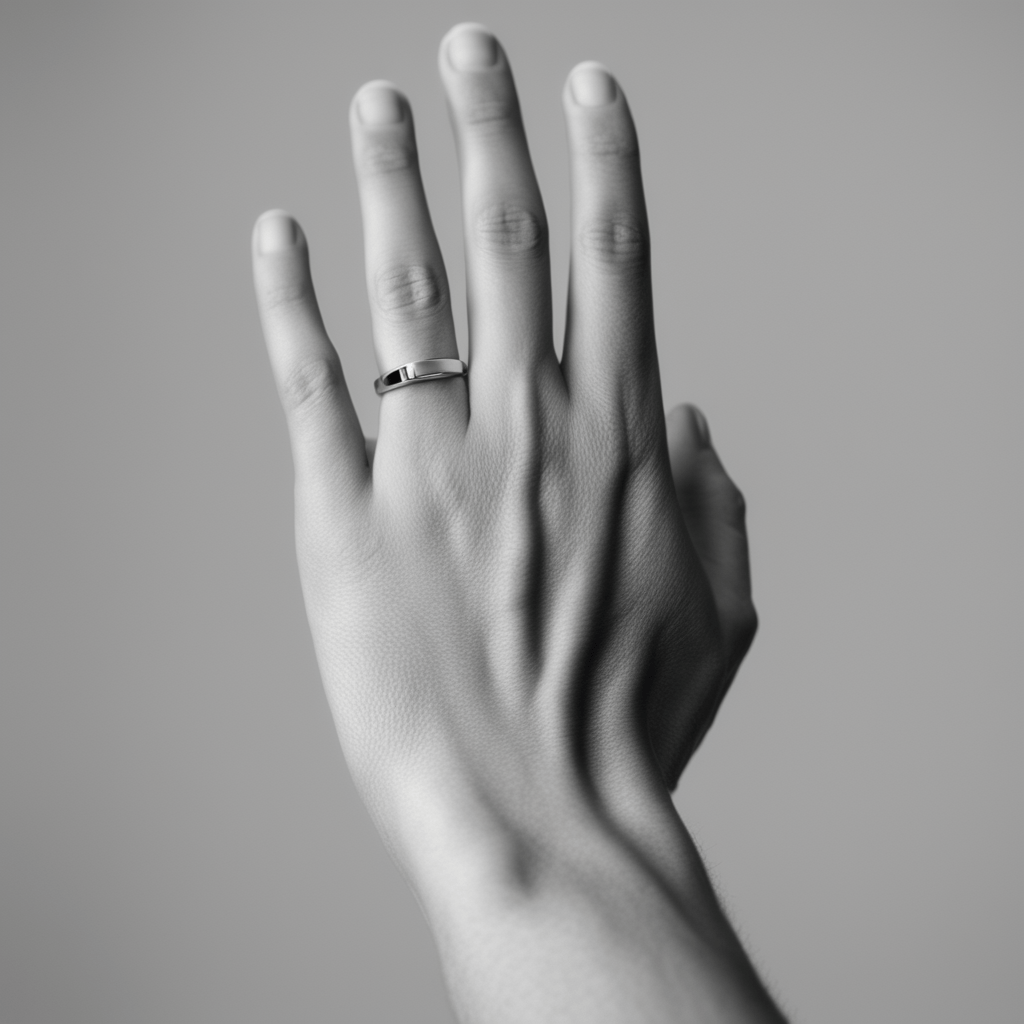}
        \caption{A high resolution, realistic image of a male hand with the ring finger adorned by a ring. The fingers are slightly apart and fully open, showcasing the ring, with a relaxed and natural posture and the thumb is slightly bent away from the palm}
    \end{subfigure}

    \begin{subfigure}[t]{0.3\linewidth}
        \includegraphics[width=\linewidth]{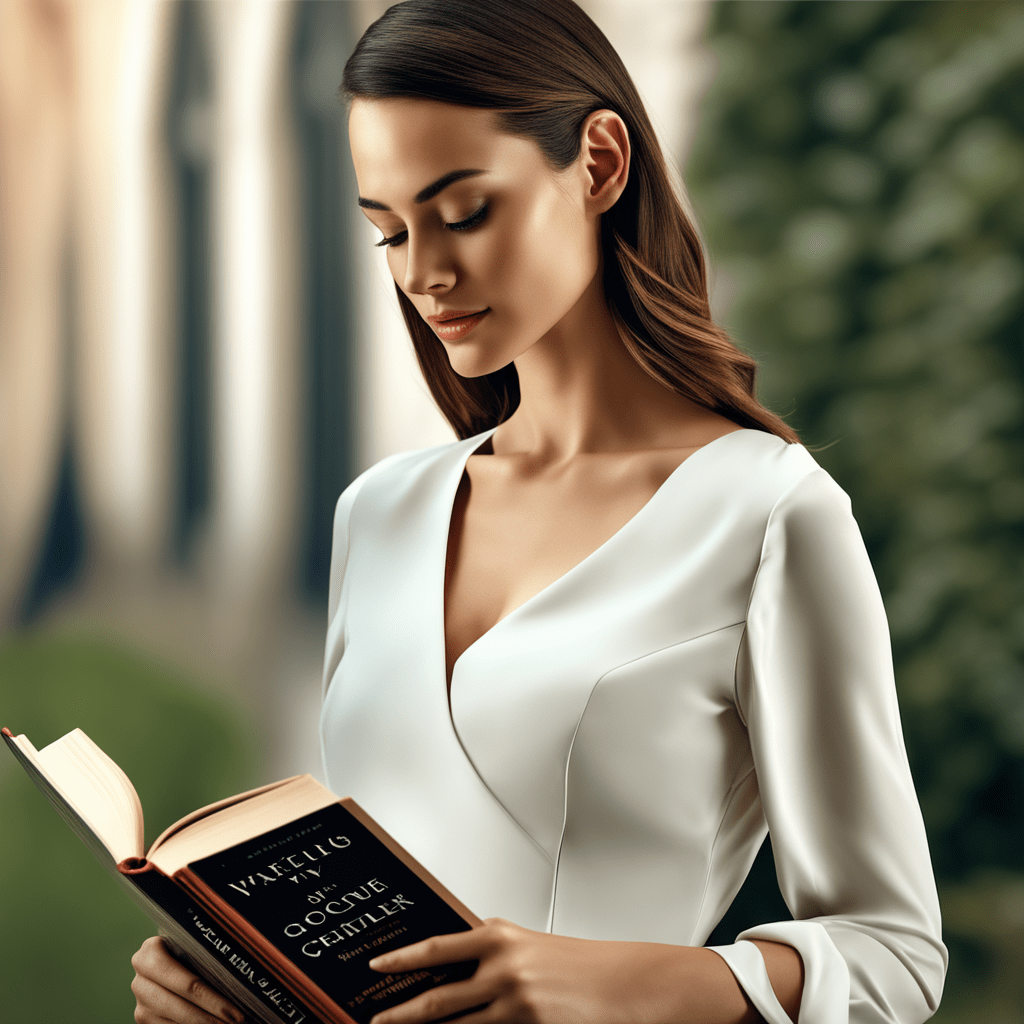}
        \caption{A woman's right hand tightly grips a large book, with the thumb, index, and middle fingers half closed around the cover, while the ring and little fingers are fully open. The left hand supports the book with all fingers half closed. (realistic, high resolution)}
    \end{subfigure}
    \begin{subfigure}[t]{0.3\linewidth}
        \includegraphics[width=\linewidth]{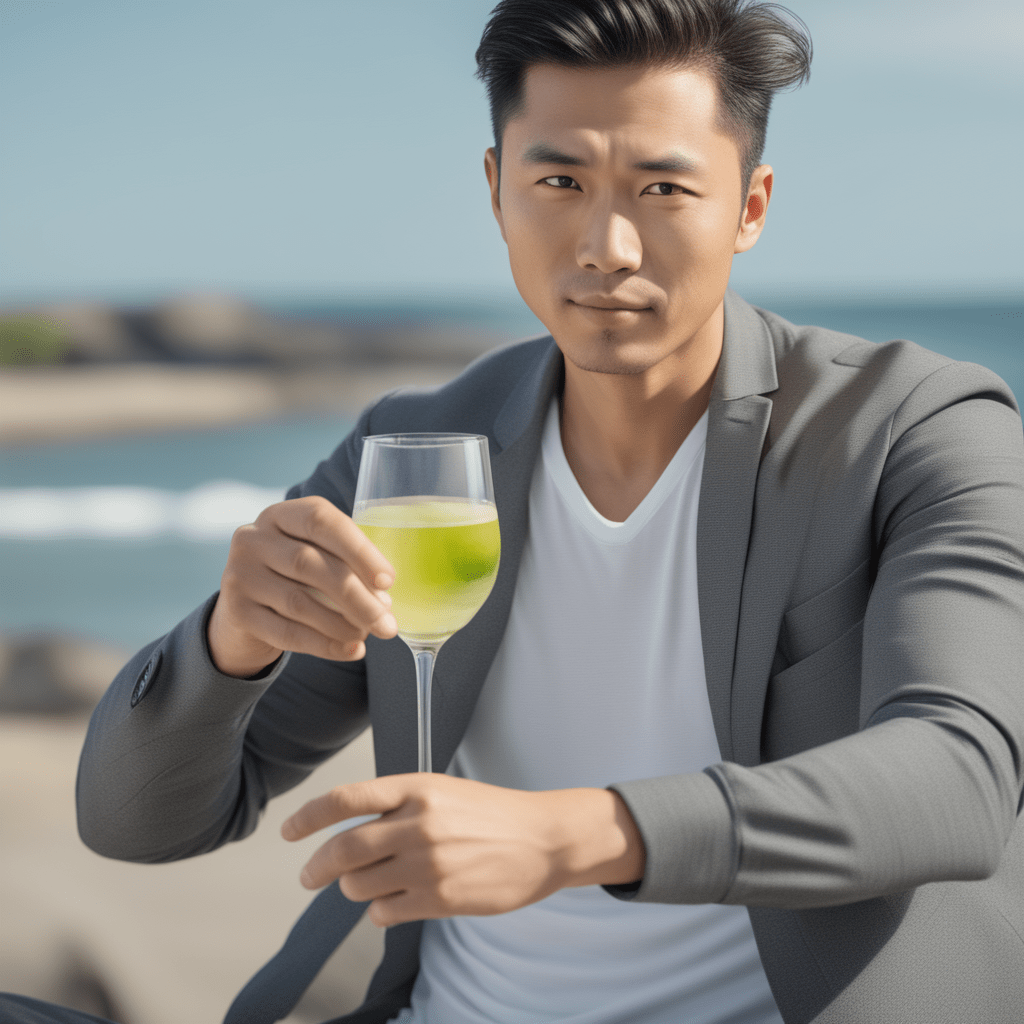}
        \caption{A chinese man holding a glass. One hand supporting and the other grasping, delicately hold a wine glass. The right hand's fingers are fully closed around the glass. The left hand provides support, with all fingers fully closed. (realistic, high resolution, detailed hand positions)}
    \end{subfigure}
    \begin{subfigure}[t]{0.3\linewidth}
        \includegraphics[width=\linewidth]{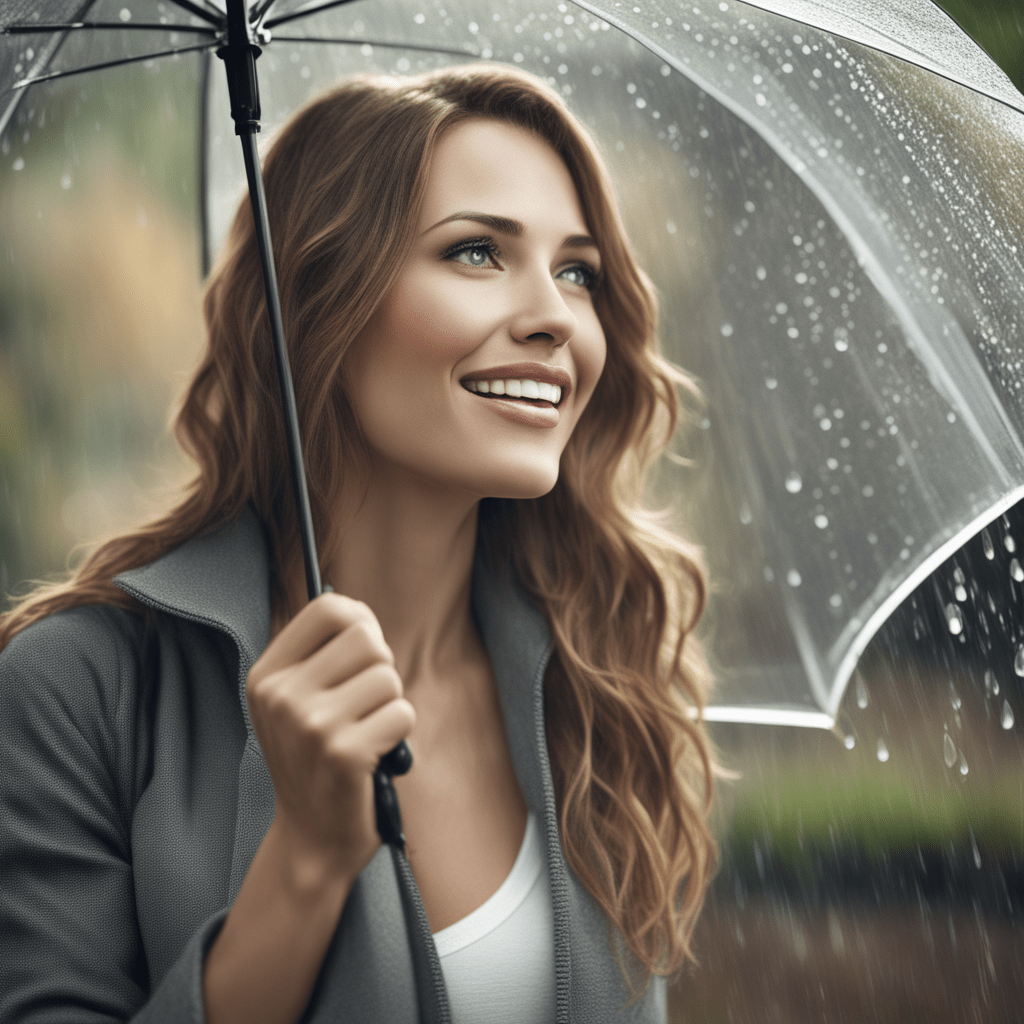}
        \caption{A high-resolution image of a female's right hand in a full hand grasp motion gracefully holding an umbrella handle, with her thumb and fingers encircling it in a fully closed position. Raindrops gently fall on the umbrella, creating a serene atmosphere}
    \end{subfigure}

    \caption{Examples of image-prompt pairs in our \shortname dataset}
    \label{images_with_prompts}
\end{figure*}

Large pre-trained vision and language models have ascended to the forefront of the ever-evolving landscape of deep learning. These models display superhuman capabilities across a spectrum of complex tasks involving image synthesis and natural language processing. The growth of these large scale models is driven in part by availability of large amounts of data to train on available from across the web \cite{COCO,visual_gen,laion5b}. Recent evolution of latent diffusion models have shown that they can also generate high-quality images guided by text prompts in a wide-variety of scenario \cite{saharia2022photorealistic}

However, despite their ability to produce photo-realistic human images, these models struggle with challenges in accurately generating specific features, such as text, teeth, and human hands (see Table-\ref{tab:failre_modes}), with human hands begin the most difficult. Generating high quality hand-object interaction images requires profound understanding of intricate hand poses and interaction. Popular datasets for training diffusion models \cite{COCO} \cite{laion5b}, lack high-quality hand-object interaction images. Consequently, current state-of-the-art diffusion models encounter limitations in their ability to generate such images.

Furthermore, it has come to light that these models exhibit biases toward certain racial and demographic groups, as highlighted in \cite{bias_danish}. This is largely attributed to the bias present in the training data. These issues can be resolved by generating high-quality synthetic data sets.

\textbf{Our contributions:} We develop \framework - a framework to generate a high-quality well-annotated synthetic Text-Image dataset. Our \framework rests on a prompt-propose-verify backbone (see fig-\ref{framework_image}). Leveraging the world knowledge stored in language models, a \textit{prompter} first generates a detailed prompt. This prompt serves as a conditioning vector to multiple \textit{proposers} to generate a set of proposed images. A \textit{verifier}, trained to accept or reject an image based on the fidelity and alignment of the prompt-image pair, then filters the pairs and augments the dataset with only the most coherent and realistic samples.

We demonstrate the effectiveness of our \framework framework by generating a high-quality text-image dataset of hands interacting with various everyday objects : \shortname. We fine-tune a StableDiffusion XL (SDXL) \cite{podell2023sdxl} model on this generated dataset. We then compare the abilities of the base SDXL \cite{podell2023sdxl} model and the one fine-tuned on the \shortname dataset on quantitative and qualitative metrics. We outperform the base model in prompts related to hand-object interactions in both quantitative (\textbf{3.32\%} and \textbf{15.9\%}  improvement over base model in CLIPScore and ImageReward respectively) and qualitative human evaluation (Avg rating \textbf{3.8} vs 2.7)\cite{clipscore,imagereward}. We also show that fine-tuning the model doesn't reduce overall generalization capability of the model by getting similar results to the base models in both measures.

\section{Related Work}
Synthetic dataset generation has emerged as a valuable technique in fields where procuring real data is arduous and unfeasible. Generative Adversarial Networks (GANs) have been proven to be invaluable for synthesizing data, particularly in classification tasks domains such as healthcare \cite{Liver_Gan} and environmental conservation \cite{environment_gan}. In addition to classification,  generative models have been used to produce pixel-level annotations \cite{DatasetGAN, BigDataset, HandsOff, IR_GAN}.

The recent advancements in diffusion models, particularly in their capacity to generate photo-realistic images \cite{saharia2022photorealistic}, has led to significant progress in enhancing classifiers through the utilization of data generated by these models \cite{synthethic_imagenet, effective_aug}. The ability to condition these model's generations on textual inputs, image editing \cite{DiffEdit, PaintByExample} and incorporating novel subjects \cite{dreambooth, text_inversion} has ushered in a new era in the realm of synthetic data, offering precise and fine-grained control over the generation process. 

Language models have exhibited remarkable capabilities when processing textual information. Despite achieving super-human performance in various tasks, these models encounter challenges in accomplishing multi-step reasoning tasks. To mitigate this limitation, prior research efforts have introduced the concept of verifiers \cite{faultaware, math_verifiers}, specialized models trained to classify inputs as either valid or invalid. Our approach aims to enhance the reliability of multi-step reasoning processes.

\section{Methodology}
\label{methods}
\begin{figure*}[hbtp]
  \centering
     \includegraphics[width=0.9\linewidth]{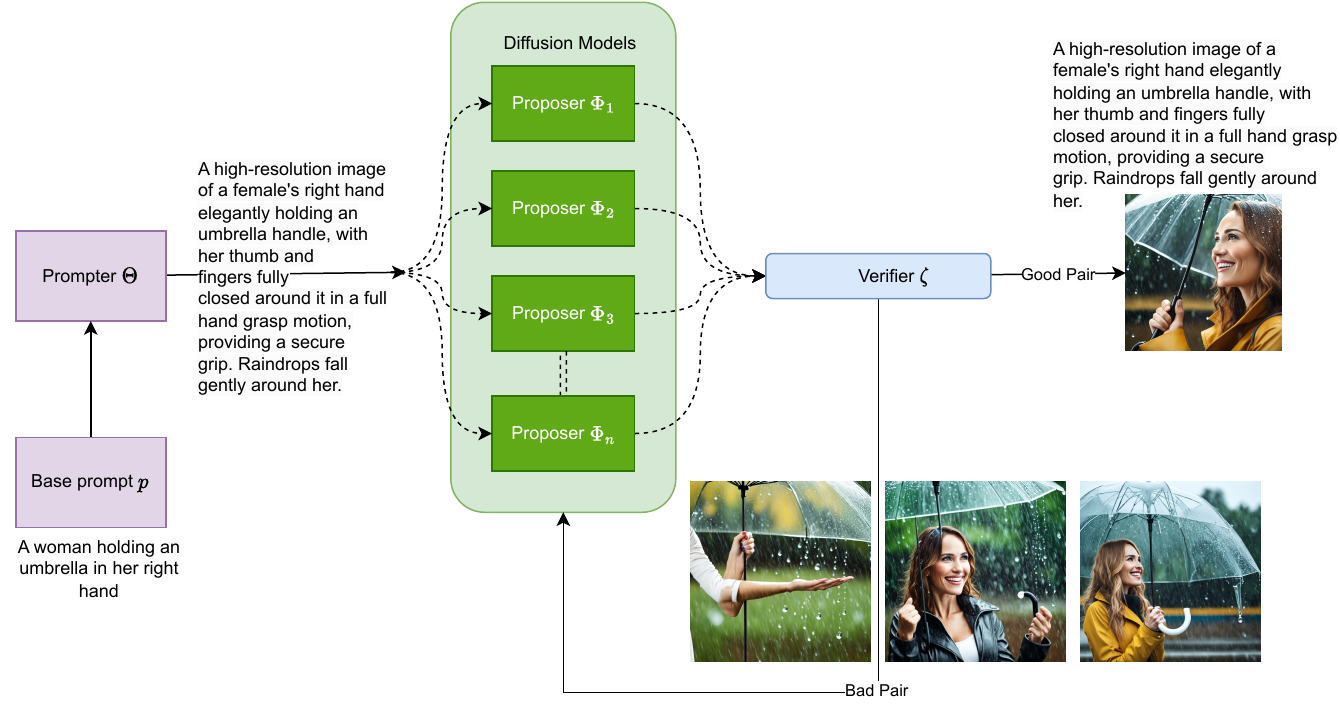}
    \caption{\textbf{Data generation pipeline of \framework}, given a base prompt $p$, the prompter $\Theta$ enriches the base prompt using the world knowledge stored in its weights. Each proposer, specialized in a unique specific object, then proposes an image $I_i$. These image-prompt pairs are then filtered using the verifier $\zeta$ based on aesthetics, accuracy, and alignment. The prompt-image pairs accepted by the verifier are added to \shortname}  
    \label{framework_image}
\end{figure*}

The proposed framework - \framework consists of three stages (see fig - \ref{framework_image})- The prompter $\Theta$, the proposers $\Phi_i(P)$ and the verifier $\zeta(I, P)$. A base prompt  $p$ is given to the prompter, in our case GPT-4, which produces a prompt $P$. The proposers, each of which is a fine-tuned diffusion model produces an image $I_i$ conditioned on input prompt vector. The image prompt pairs $<I,P>$ are then fed into the verifier, parameterized by $\zeta$, that predicts whether the image is accurate and aligns well with the prompt. 

\subsection{Prompter $\Theta$}

Datasets like COCO \cite{COCO} and LAION-5B\cite{laion5b} lack intricate pose, structural, and interaction information pertaining to entities like hands. Diffusion models trained on these datasets \cite{sd_original} face challenges in understanding the intricacies of such entities. To address this challenge, we propose to use a prompter to augment \shortname by introducing detailed textual prompts. The proposer is designed to generate prompts that accentuate the anatomical aspects of hands and the relative spatial positions of hand features concerning objects they interact with.

Pre-trained large language models (LLMs) are trained on extensive internet corpora, and thus encapsulate a wide spectrum of worldly knowledge.This inherent attribute renders them suitable to be used as \textit{prompters} $\Theta$, due to their capability to furnish intricate and comprehensive object descriptions. Specifically, we use GPT-4 to generate detailed prompts $P$, based on a given base prompt $p$. To ensure that \shortname covers a wide range of object categories, we prompt GPT-4 to produce texts describing hand-object interactions across a variety of object classes. This acts as the base prompt $p$. Refer to Table-\ref{breakup-table} in the appendix for an overview of the distribution of object classes.

\textbf{Handling possible noise from Language Models}
Language models quite often hallucinate \cite{hallucination1, hallucination2, hallucination3}, this can introduce noise in diffusion models and worsen their generations. To avoid this, we propose to introduce a deterministic component. We define a Domain-Specific-Language (DSL), where each hand position, object position and their interaction can be expressed as a code in DSL. Given the DSL program, we run it through a rule-based programs that tells if the position of each finger, palm, object interaction is physically possible given the object shape and size. We then give this program to GPT-4 to refine it into a language prompt. This step ensures that none of the prompts contain hand positions that are not physically possible. Example of one such DSL program is given in the appendix \ref{dsl-prog-example}

\textbf{Ensuring fairness of the generated dataset}
Previous studies have demonstrated the existence of bias and disproportionate representation in large-scale diffusion models \cite{bias_stable,bias_danish}. Deep-learning models tend to amplify bias present in the training data \cite{systematic_bias}, to mitigate this issue we ensure that \shortname represents populations from different races, ethnicities, and genders appropriately. We explicitly model the Prompter $\Theta$ to generate a versatile array of prompts. The distribution of ethnicities within the dataset is given in Table~\ref{breakup-table}. The meta prompt given to GPT-4 to generate input for diffusion models is given in the supplementary.

\subsection{Proposer $\Phi$}
\label{dream-intro}
Diffusion models learn the distribution of data by denoising in a stepwise manner. Given an initial noise map $\epsilon \in \mathcal{N}(0, \mathbf{I})$ conditioned on an input context vector $c = \tau(P)$ where $\tau$ is a text encoder and $P$ is the text prompt. Attributing to 
a lack of data, these models face challenges in generalizing hand images to different and complex scenarios. To curate a dataset with good hand images, it becomes imperative for the generator model to be able to produce realistic hand images with human-like anatomy. One natural choice is to finetune a diffusion model to generate good hands. However, given the scarcity of well-annotated hand datasets of fine quality, we decided to leverage DreamBooth\cite{dreambooth} to finetune a diffusion model in a few-shot manner. DreamBooth\cite{dreambooth} implants a novel subject into the output space of diffusion models using a unique identifier for the subject.

Hands have multiple degrees of freedom and limitless granularity of movement. To ensure the fidelity of images being generated we fine-tune multiple \textit{proposer} $\Phi$ models using DreamBooth \cite{dreambooth}. Inspired by the classifications made by \cite{jian2023affordpose}, we first classify hand-object interaction into categories based on hand poses. We then fine-tune diffusion models using DreamBooth\cite{dreambooth} for each category. At inference time, given a text prompt $P$, we use these \textit{proposers} to generate images $I_i$. Each of these paired text-image entities undergoes evaluation by the verifier. Any pair accepted by the verifier is then incorporated into the dataset. If all pairs are rejected by the verifier, the proposers are again prompted to generate fresh images. 

\subsection{Verifier $\zeta$}

Although specialized proposers generate high-fidelity images given the prompt, they fail to generalize to novel scenarios (see Fig-\ref{poor-examples}) attributing to the complex nature of hand-object interaction. This invokes the need for a robust verifier that ensures that only high-quality and aligned image-prompt pairs are added to the data set.

The choice of the verifier architecture depends upon the task at hand, owing to the complex nature of hand images, we finetune a ViLT \cite{vilt} based verifier, that is trained on a subset of 5000 manually annotated text-images pairs generated by the proposers finetuned using DreamBooth \cite{dreambooth}.   

\begin{table*}[]
\centering
\begin{tabular}{ccccccc}\\
\toprule
\multicolumn{1}{l}{} & \multicolumn{2}{c}{Quantitative} & \multicolumn{3}{c}{Qualitative} \\
Model Name & CLIPScore $\uparrow$ & ImageReward* $\uparrow$ & Fidelity & Alignment & Overall\\
\hline
\textbf{Hand Images Prompts }\\
\hline
Base &     31.64      &  0.44 & 2.60 & 2.66 & 2.70 \\
DreamBooth &    32.04   &  0.38 & 2.86 & 2.80 & 2.70\\
Finetuned \textit{(\textbf{Ours})}    &   \textbf{32.69} & \textbf{0.51} & \textbf{3.73} & \textbf{3.73} & \textbf{3.8}        \\
w/o Prompter (Ablation) & 28.51 & 0.18 & - & - & - \\
w/o Proposer (Ablation)& 31.54& 0.50 & - & - & - \\
w/o Verifier (Ablation)& 30.90& 0.46 & - & - & - \\
Generic hand dataset (Ablation)& 31.60 & 0.35 & - & - & - \\
\hline
\textbf{DrawBench Prompts}\\
\hline
Base &           \textbf{28.00}  & 0.61 & 4.43 & 4.13 & \textbf{4.44}\\
Finetuned \textit{(Ours)}      &      27.97 & \textbf{0.64} & \textbf{4.46} & \textbf{4.26} & 4.43   \\
\bottomrule

\end{tabular}
\caption{Comparison of model finetuned on \shortname with base stable diffusion model and DreamBooth finetuned model. * ImageReward scores are normalized to 0.0 to 1.0 using minimum and maximum values of ImageReward for a given model. Qualitative results for ablations are not shown due to resource unavailability.}
\label{score-table-human}
\end{table*}

\section{Experimental Results}

In this section we present qualitative and quantitative results on the downstream task performance using the \shortname dataset using our \framework framework. A glimpse of the images generated by the finetuned model can be seen in the supplementary material. Detailed implementation details can be found in Appendix \ref{implement_details} 
\subsection{Finetuning model using \shortname data}
\label{fine-tuning-exp}
We generate $\sim$ 10k high-quality annotated text-image pairs using \framework as described in Sec-\ref{methods} to curate the dataset - \shortname. We then finetune the Stable Diffusion XL \cite{podell2023sdxl} model on \shortname using LoRA \cite{lora}.

We compare the performance of finetuned versus base Stable Diffusion XL \cite{podell2023sdxl} and DreamBooth finetuned models using the CLIPScore \cite{clipscore} and normalized ImageReward \cite{imagereward}, we observe that (see Table-\ref{score-table-human}), the CLIPScore for finetuned model is the highest out of the three models and outperforms the Base model by \textbf{\textit{3.32\%}} and DreamBooth ensembles by  \textbf{\textit{2.03\%}}. Similarly, for the ImageReward metric, we outperform the Base model by \textbf{\textit{15.90\%}} and DreamBooth ensembles by  \textbf{\textit{34.21\%}}.

\begin{table*}[hbtp]
    \centering
    \begin{tabular}{|m{6cm}|ccc|}
    \hline
         \textbf{Detailed input Prompts} &  \textbf{Base} & \textbf{DreamBooth} & \textbf{Finetuned} \\
         \hline
         \begin{minipage}{\linewidth}
         \vspace{-1em}
             An Asian female holding a drink glass. Right hand wraps around the base of a small glass, with all fingers half closed around the object. The left hand supports the glass with all fingers half closed. The glass is in full contact with the right hand's fingers. (realistic, high resolution, clear hand anatomy)
        \end{minipage}
        &
        \begin{minipage}{0.15\linewidth} \includegraphics[width=\textwidth]{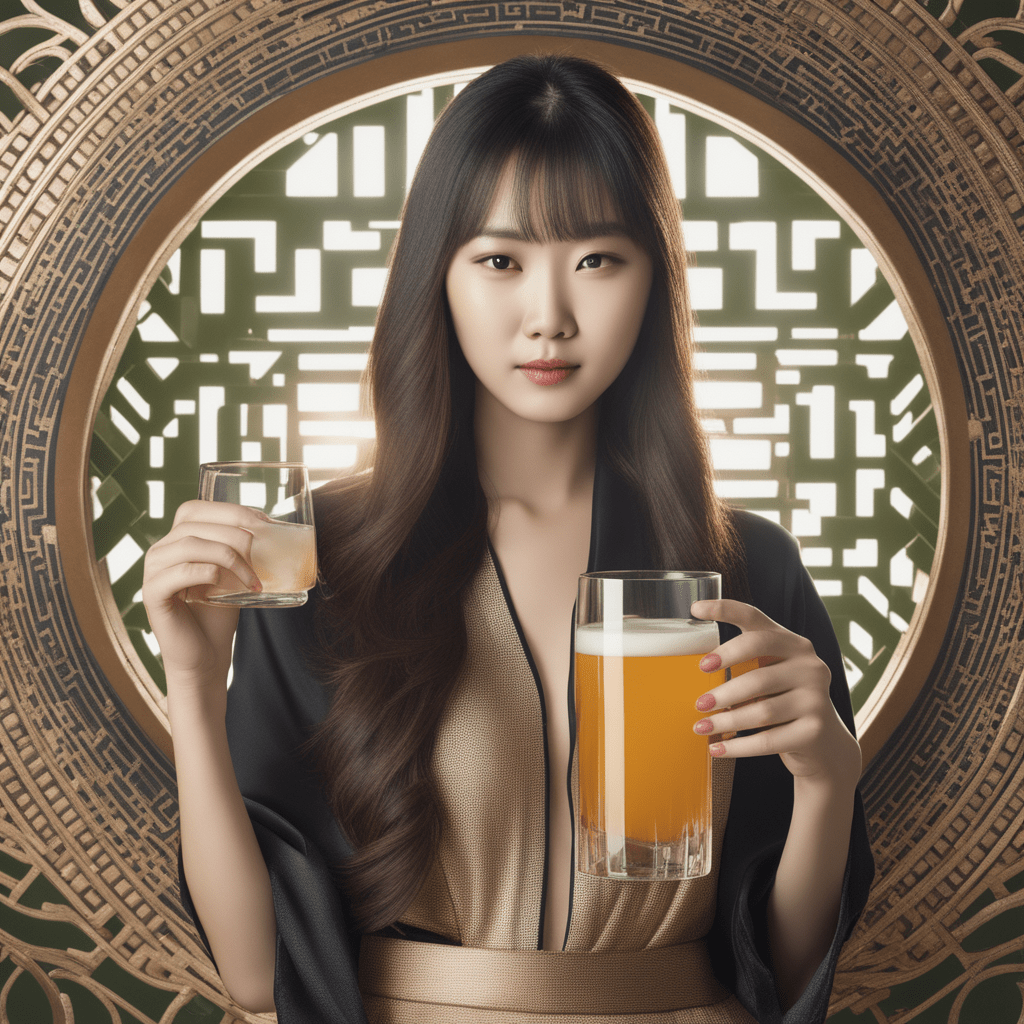} \end{minipage} & 
        \begin{minipage}{0.15\linewidth} \includegraphics[width=\textwidth]{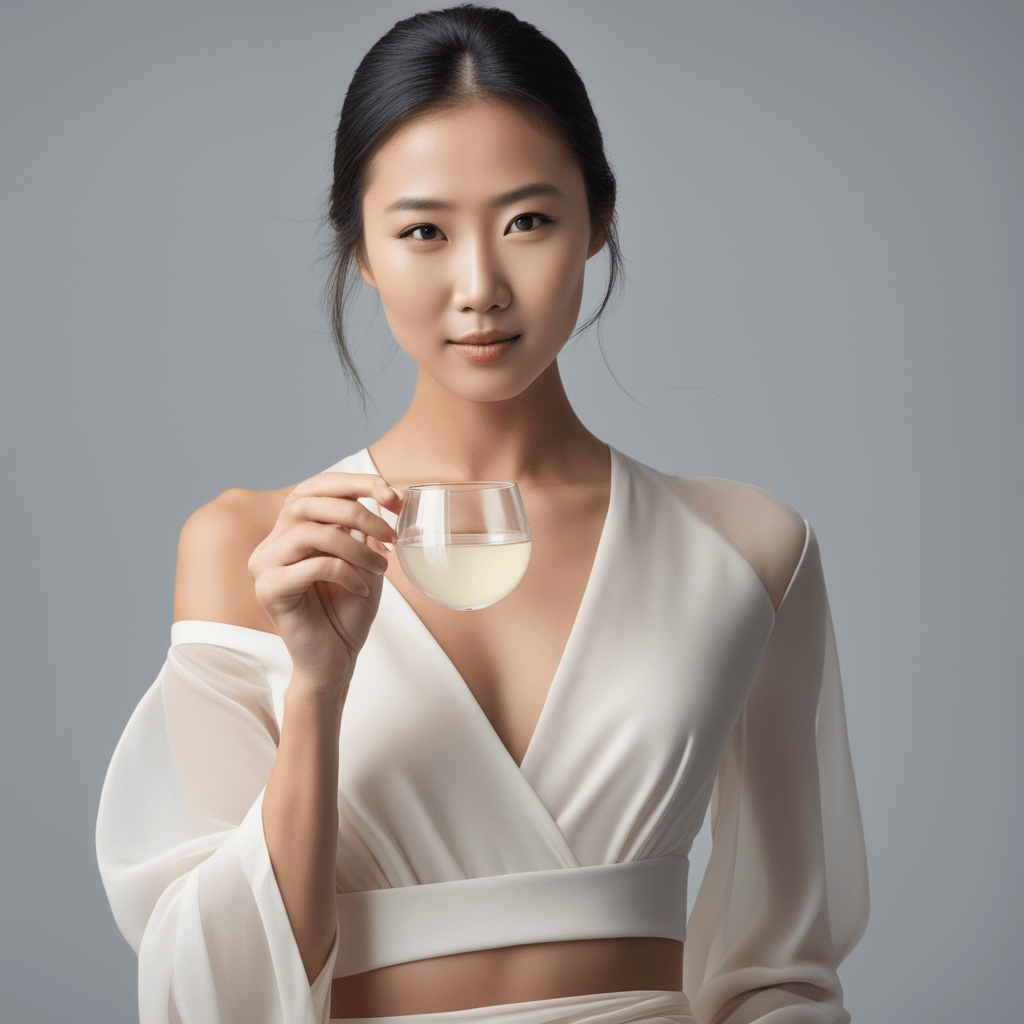} \end{minipage}&  
        \begin{minipage}{0.15\linewidth} \includegraphics[width=\textwidth]{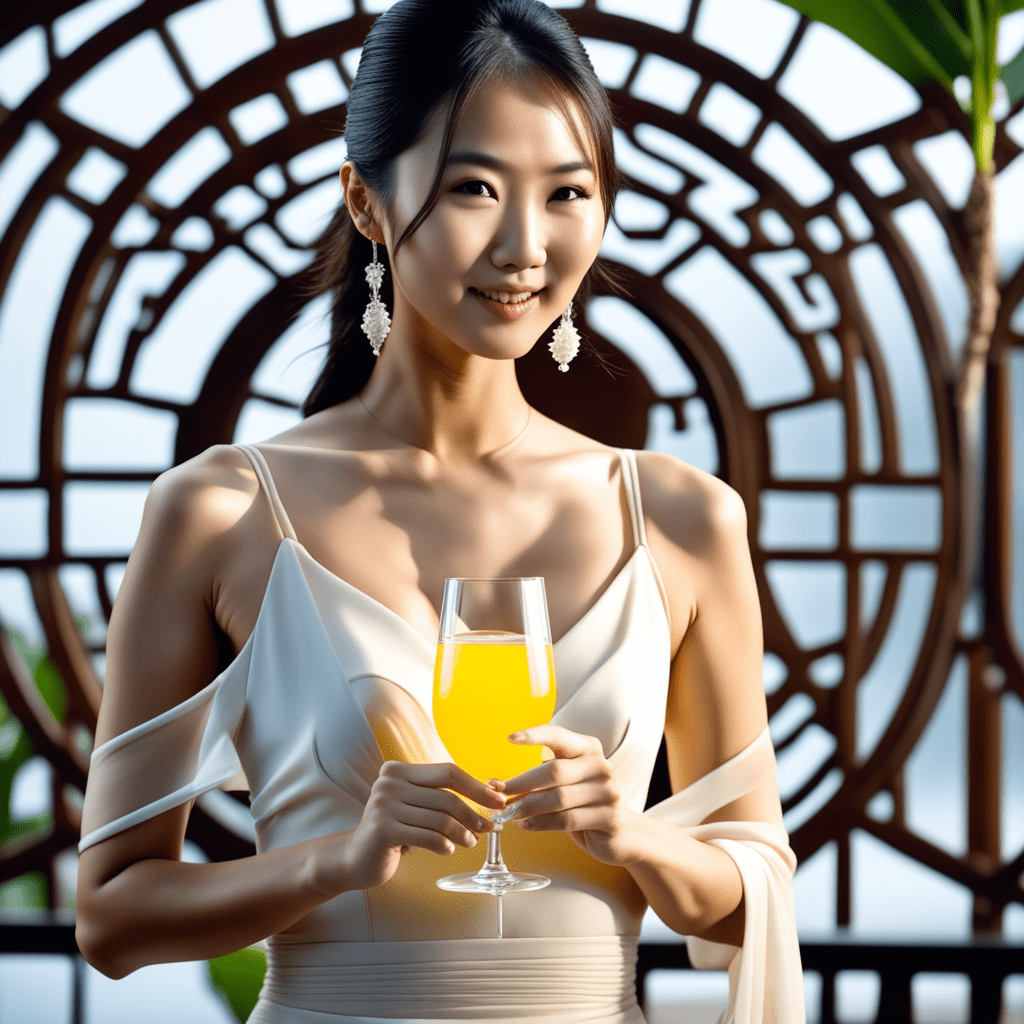} \end{minipage}\\
\hline
        \begin{minipage}{\linewidth}
        \vspace{-1em}
             A realistic 4k image of an asian man's right hand fully wrapping an umbrella, with the thumb half closed and the fingers fully closed around the object. The left hand is not visible. High resolution details of hand and object.\\
        \end{minipage}
        &
        \begin{minipage}{0.15\linewidth} \includegraphics[width=\textwidth]{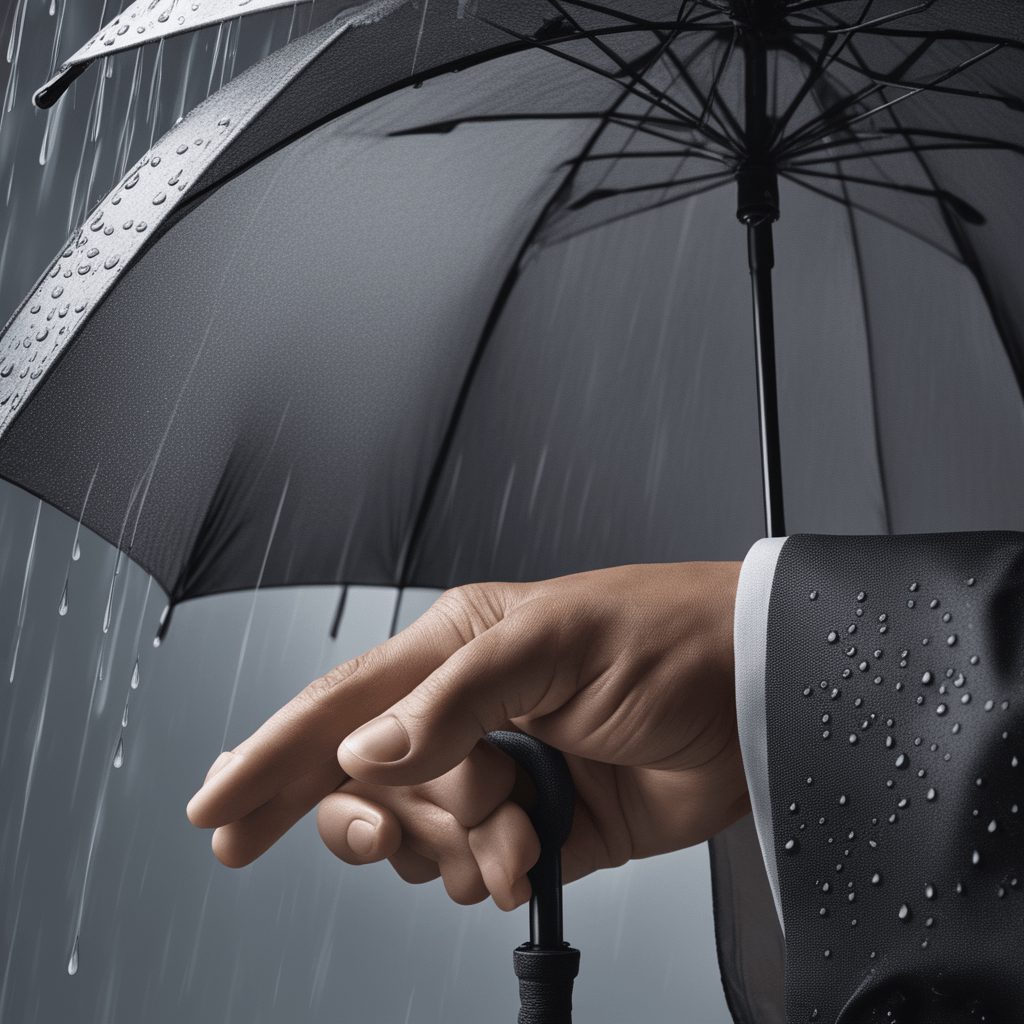} \end{minipage} & 
        \begin{minipage}{0.15\linewidth} \includegraphics[width=\textwidth]{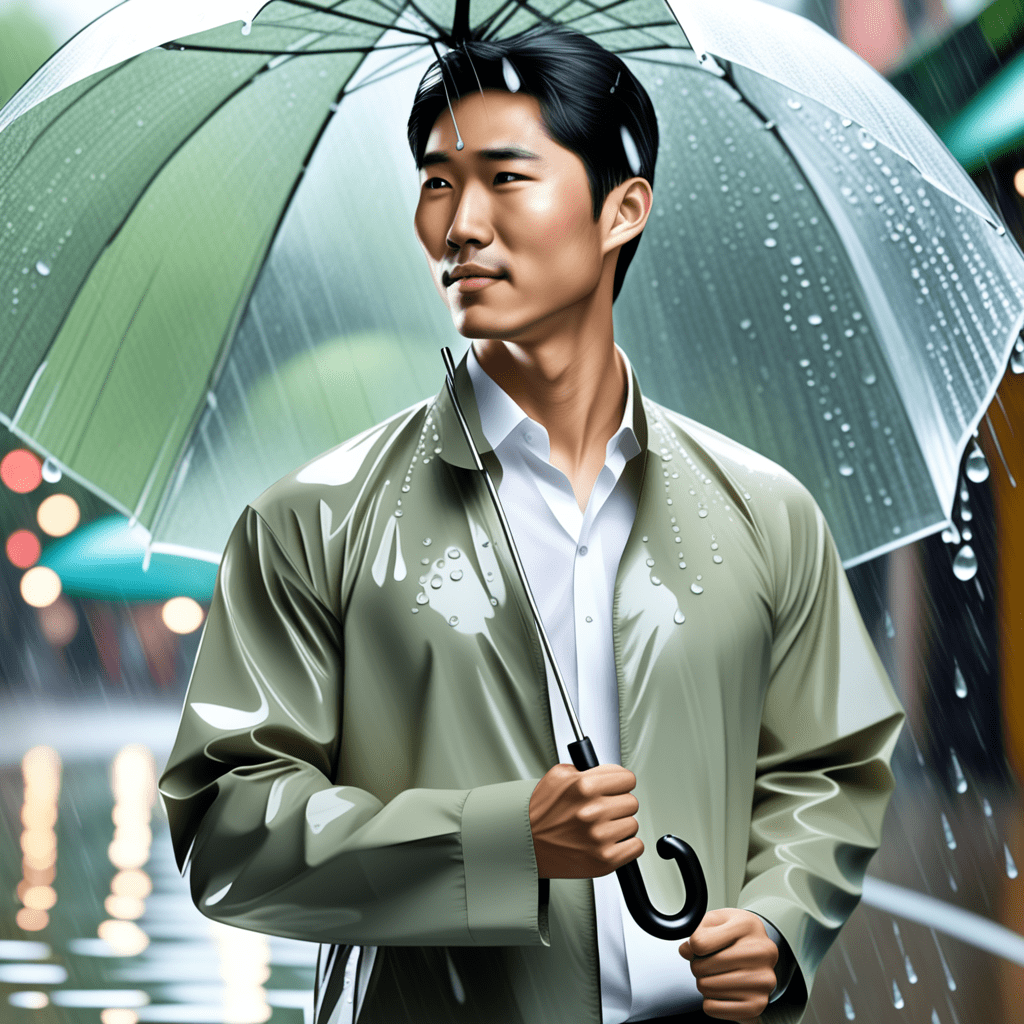} \end{minipage}&  
        \begin{minipage}{0.15\linewidth} \includegraphics[width=\textwidth]{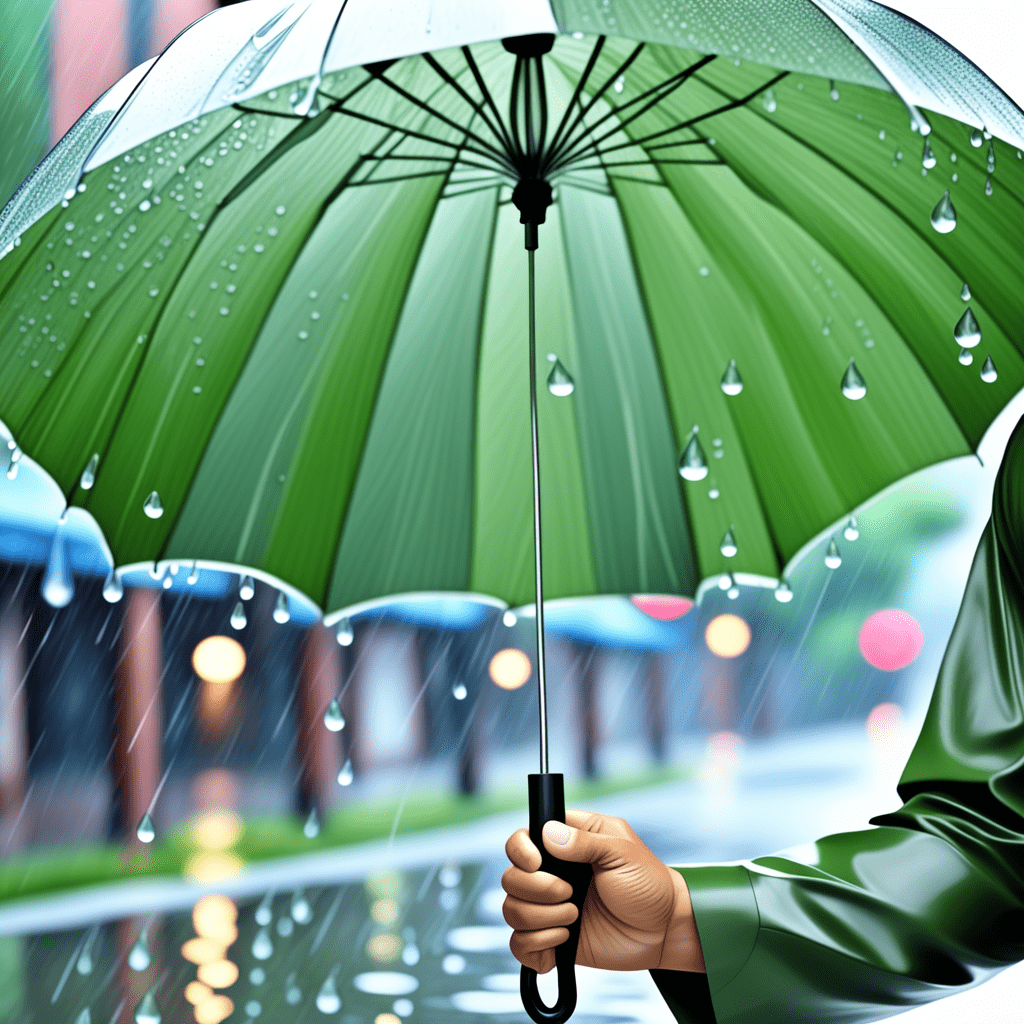} \end{minipage}\\
\hline
        \begin{minipage}{\linewidth}
        \vspace{-1em}
             A realistic, high-resolution image of a tanned female right hand in a supportive gesture, with all fingers fully-open. The tiny ring is positioned at the base of the middle finger, without touching the palm or any other fingers. Rested on a satin cloth. The fabric appears smooth and luxurious.
        \end{minipage}
        &
        \begin{minipage}{0.15\linewidth} \includegraphics[width=\textwidth]{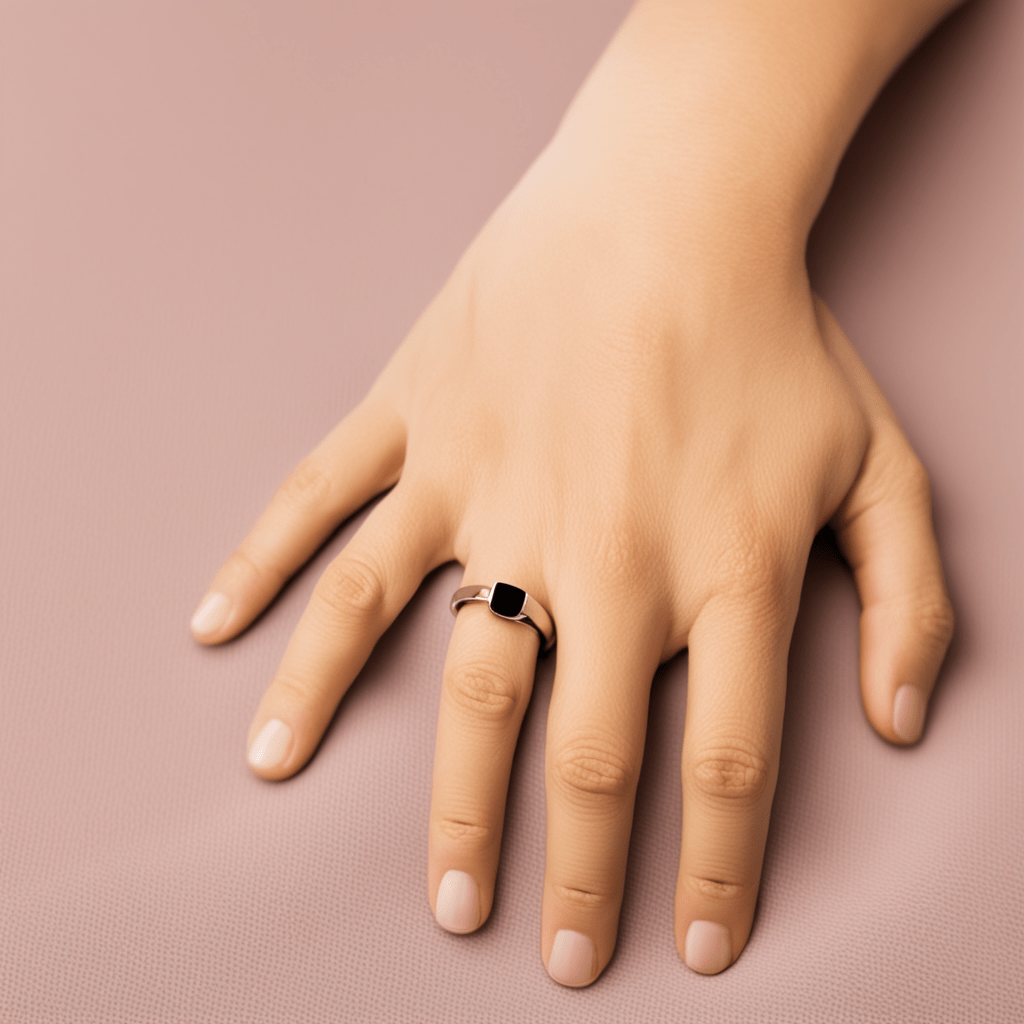} \end{minipage} & 
        \begin{minipage}{0.15\linewidth} \includegraphics[width=\textwidth]{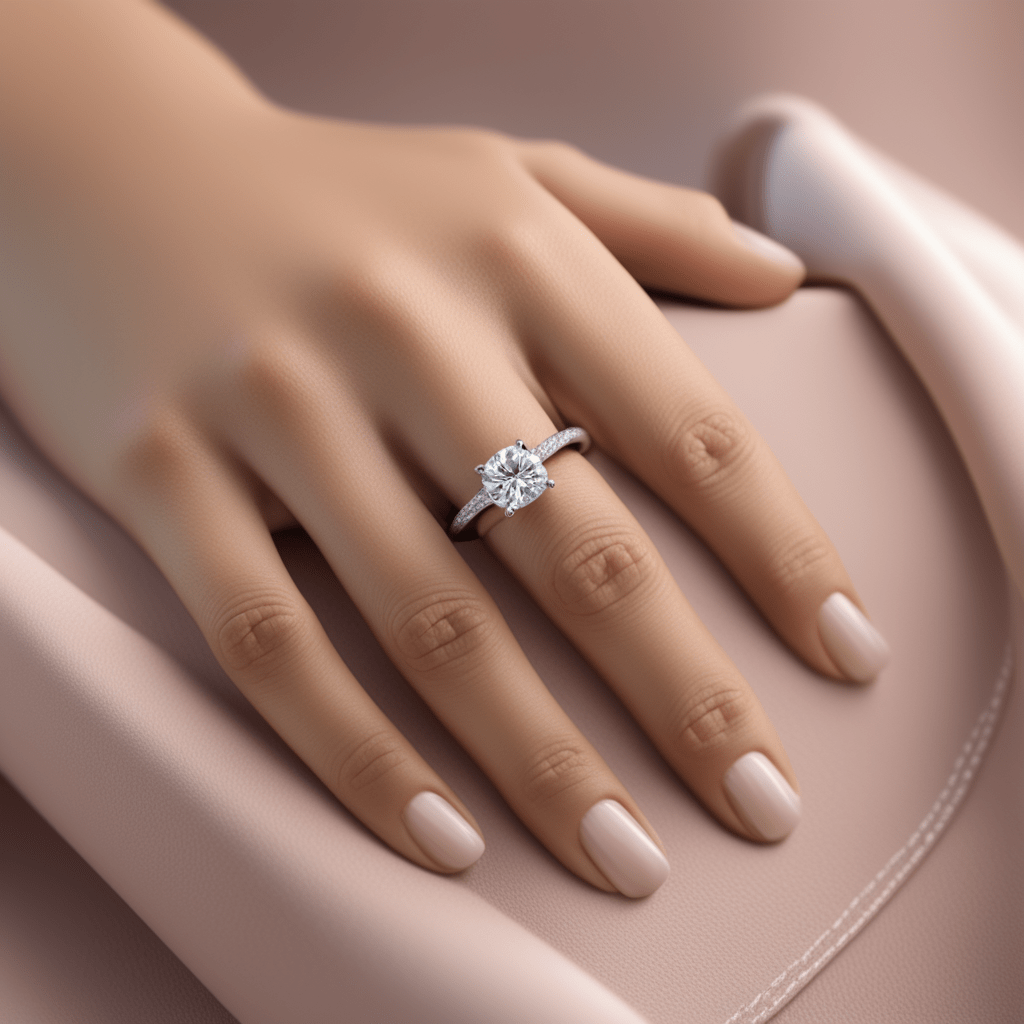} \end{minipage}&  
        \begin{minipage}{0.15\linewidth} \includegraphics[width=\textwidth]{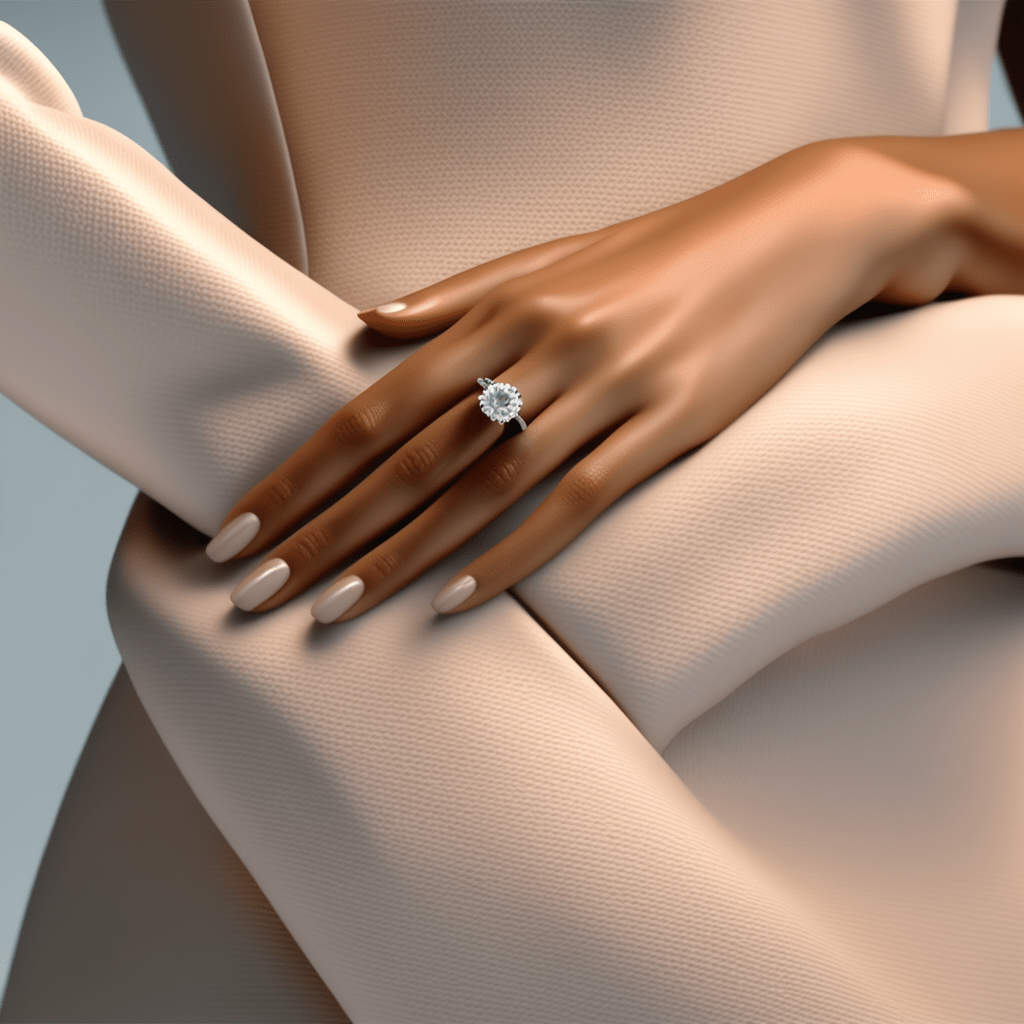} \end{minipage}\\
      \hline
      
    \end{tabular}
    \caption{Comparision of model fine-tuned using \shortname with DreamBooth finetuned and base stable diffusion models. }
    \label{tab:comparison}
\end{table*}

Deep Learning models are susceptible to catastrophic forgetting when trained on fine-grained concepts \cite{cat_forget}.To assess the impact of fine-tuning the base model using \shortname on its performance when processing prompts involving objects other than hands, we conduct a comparative analysis of its CLIPScore and ImageReward against the base model on the \textit{DrawBench}\cite{drawbench} dataset . Table-\ref{score-table-human} shows training on \shortname does not yield a significant reduction in CLIPScore or ImageReward thereby indicating that the model does not suffer from catastrophic forgetting or language drift.

\subsection{Human study on images generated by finetuned model}
The CLIPScore and ImageReward metrics we used are trained on text-image pairs with limited amount and generic data points, exhibit limitations in effectively distinguishing between good-quality and bad hand image generations. To provide a more robust assessment of the model's performance after fine-tuning on the \shortname dataset, we conducted a qualitative human study involving five participants excluding the authors. Each participant evaluated a total of 30 images across three key dimensions: Fidelity, Prompt Alignment, and an overall rating on a scale ranging from 1 (indicating poor) to 5 (indicating excellent). Of the 12 prompts used, more half included hand-object interactions, while the remainder were randomly selected from the DrawBench dataset. Detailed information regarding the user study is available in Section \ref{user_study_example}. The results of the human evaluation, as depicted in Table \ref{score-table-human}, consistently favored the fine-tuned model across all three dimensions in comparison to both the base model and the DreamBooth ensembles. In the terms of Fidelity the finetuned model (\textbf{3.73}) outperforms the Base model (2.60) and the DreamBooth ensembles (2.86). In the prompt-alignment dimension the finetuned model (\textbf{3.73}) outperforms the Base models (2.66) and the DreamBooth ensembles (2.80). Similarly in the overall score the finetuned model (3.8) outperforms the Base model (2.7) and the DreamBooth ensembles (2.7). The model fine-tuned on the \shortname dataset exhibited comparable performance to the base model for general prompts sourced from DrawBench.

\subsection{Ablation Study}

To prove the effectiveness of our proposed \framework framework, we perform an abalation study. To show the effectiveness of each individual component in the pipeline we generate 3 datasets and fine-tune 3 SDXL models as discussed in Sec \ref{fine-tuning-exp}.

To perform the ablation study we generate these 3 datasets, where we remove one out of the 3 components and generate text-image pairs :  
\begin{enumerate}
    \item \textbf{Without Prompter}: We fine-tune a model using simple just the base prompt $p$ not enriched by the prompter
    \item \textbf{Without Proposers}: We fine-tune a model using a dataset where the images are not generated by specialized DreamBooth fine-tuned diffusion Proposers.
    \item \textbf{Without Verifier}: We fine-tune a model using a dataset where the images generated are not `verified' by the Verifier $\zeta$
\end{enumerate}

We compute the CLIPScore and ImageReward metrics for all these 3 models and the results are shown in \ref{score-table-human}. We show a significant increase in both the metrics when all 3 components of our framework are used instead of just using 1 particular component.

To show the overall effectiveness of our framework we also fine-tune a diffusion model on a dataset of real hand images however with the non-detailed prompt `hands'. The model fine-tuned using the \shortname dataset outperforms the generic dataset one. More details on the generic dataset can be found in the Appendix.  

\section{Discussion and Future Work}
In this work, we introduce \framework, a novel approach that streamlines the creation of high-quality, fair, and detailed Text-Image synthetic datasets, circumventing the labor-intensive manual annotation processes associated with traditional datasets like \cite{COCO,visual_gen} or the extraction of unstructured data from the internet \cite{laion5b}. Our framework's efficacy is demonstrated through the generation of a detailed Text-Image dataset focused on hand-object interaction.We finetune a SDXL model on this dataset resulting in notable improvements over the base model across both qualitative and quantitative evaluation metrics. Whilst we focused on text-to-image Diffusion models the generated Text-Image data can be used to improve performance of image captioning and other multi-modal models.

Looking ahead, interesting questions arise: Can we leverage verifier feedback to refine input prompts or further fine-tune proposer models? Moreover, when extended to accommodate 3D data, could \framework be employed to generate supervised data for training for robotic hand-object manipulation tasks? These avenues hold promise for future exploration.

\appendix
\section*{Appendix}
\section{Implementation Details}
\label{implement_details}

\textbf{Fine-tuning SDXL using DreamBooth} We fine-tune multiple SDXL models using DreamBooth specialized on generating hands in different poses like holding an object, grasping, open palm, having accessories, making hand gestures etc. We source 5 images from unsplash.com for each pose type and train the model for 500 training steps with an instance prompt and the unique identifier for the class being [sks].Training the DreamBooth model used $\sim$17 GB of VRAM. We decided to use the SDXL Base 1.0 from StabilityAI as the base model. Instead of fine-tuning a SDXL model using DreamBooth we can use other Diffusion models like DALL.E2, MidJourney and other fine-tuned variants of SDXL which have a high class prior and generate good quality images for that class. We had to fine-tune the proposer due to very poor prior hand generation capabilities of the model. See \ref{distribution_images} for more.

\textbf{Prompter $\phi$} We use GPT-4 to generate high quality positive and negative prompts for the DreamBooth tuned model using few-shot learning. We provide a few examples to GPT of good positive and negative prompts using In-Context learning and ask GPT to come up with a base prompt of a hand interacting with everyday objects and refine it. The prompts are then routed to the approriate DreamBooth model based on the content of the prompt.

\textbf{Training of the Verifier $\zeta$} We generated $\sim$2000 images using the Prompter-Proposer pipeline. Then these images were labelled into Binary classes [Good/Bad] based on the Prompt-Alignment and the quality of the generated hands. We then used a pre-trained ViLT model and add a fully connected layer to network acting as the final output logit classifying each $<I_i,P>$ as either good or bad. The training data was processed using the ViLT processor for the respective model and the model was trained for 10 Epochs with the Adam optimizer with a LR scheduler. Other verifiers can be explored depending upon the dataset being generated, CLIPScore might be a good filter if the CLIP model embeddings are sensitive to the data to filter out the bad images. Simple CNN based classifiers might also suffice for single pose use-cases. Multiple different verifiers can also be used in cases where we have multiple classes and poses, each proposer could feed to it's verifier.  

\textbf{Fine-tuning SDXL on \shortname using LoRA} The SDXL Base model's U-Net and text encoders (SDXL has 2 text encoders) were finetuned on the \shortname dataset for 10 epochs using LoRA \cite{lora}. We used an initial lr=$10^{-4}$ and batch size = 4 and a cosine lr scheduler. The training process took around $\sim$ 9 hours to complete on an A100 GPU and consumed $\sim$ 64GB of VRAM.

\textbf{Image generation using SDXL} We use HuggingFace Diffusers' diffusion pipeline to do inference on the diffusion models. We first feed the prompt to the base SDXL model (or the LoRA \cite{lora} adapted DreamBooth/Finetuned model) and generate the latents. These latents are then fed to the SDXL 1.0 Refiner along with the prompt to generate the final image. We do 100 steps of denoising : 80 for the base model and 20 for the refiner. We follow a guidance scale of 7 and use the DPMSolverMultistepScheduler. 

\section{Distribution of fingers in base SDXL model}
\label{distribution_images} We generate 100 different images of hands in various different poses and interacting with different objects using the default 1.0 SDXL model and then the images were manually assigned into bins depending on the perceived number of fingers visible on the screen, another category for bad generation was also added for when the fingers were clearly not visible. Out of the 100 images 97 of them had visible and distinct fingers. We then plotted an histogram to get an idea about the observed frequency of the number of fingers and saw that 6 fingers was the most frequent generation class of the SDXL model.

\section{User study details}
\label{user_study_example}
\textbf{Definition of the dimensions} We used 3 dimensions to rate the generated images : All these dimensions were rated from 1-5 where 1 was Bad and 5 was excellent. \\
\textbf{Fidelity} : How good does the generated image look ? Does it look aesthetically pleasing ? Is the hand-object interaction correct and images look grounded in reality ? \\
\textbf{Prompt-Alignment}: Does the generated image follow what was asked in the prompt ?\\ 
\textbf{Overall Score} : Rating of the image based on overall quality and look. 

We randomly selected 6 different prompts and generated 1 image each from 3 Models (Base, DreamBooth, FT-SDXL).We also selected 6 Random prompts from the DrawBench prompts list \cite{drawbench}{DrawBench} prompts list and generated 1 image each from the 2 Models (Base, FT-SDXL). Then a form was created using Microsoft Forms which had the Image, Prompt and the rating areas across the three dimensions. The order of the questions was randomized for each participant and the participant did not source model of the generated images. We then computed the average rating across each dimension for each of the groups by model and prompt type. 

\begin{table*}[h]
\centering
\begin{tabular}{lllllll}
    \toprule
            & \multicolumn{5}{c}{Race}         &       \\
    Object Type & Light skinned & Dark Skinned & Asian & Indian &  Latin & Total \\
    \midrule
    Kitchen objects  & 100 & 200 & 200 & 200 & 200 & 900        \\
    Sports Objects   & 100 & 100 & 200 & 200 & 200 & 800    \\
    Electronics      & 100 & 100 & 200 & 200 & 200 & 800\\
    Musical Instruments & 200 & 200 & 200 & 200 & 300 & 1100        \\
    Hardware tools & 200 & 200 & 200 & 200 & 200 & 1000\\
    Art supplies & 100 & 100 & 100 & 100 & 100 & 500\\
    Medical Instruments & 100 & 100 & 100 & 100 & 100 & 500\\
    Gardening tools & 100 & 100 & 100 & 100 & 100 & 500\\
    Vehicle Interior & 100 & 100 & 100 & 100 & 100 & 500\\
    Straight hand & 100 & 200 & 300 & 200 & 200 & 1000 \\
    Household supplies & 100 & 100 & 100 & 100 & 100 & 500\\
    Office supplies & 100 & 100 & 100 & 100 & 100 & 500\\
    Miscellaneous & 300 & 300 & 300 & 300 & 300 & 1500\\
    \midrule
    Total & 1700 & 1900 & 2300 & 2100 & 2100 & 10100\\
    \bottomrule
\end{tabular}
\caption{Breakup of the generated \shortname dataset by Race and Object type}
\label{breakup-table}
\end{table*}

\begin{figure}[h]
\centering
\begin{tabular}{cccc}
\includegraphics[width = 0.2\columnwidth]{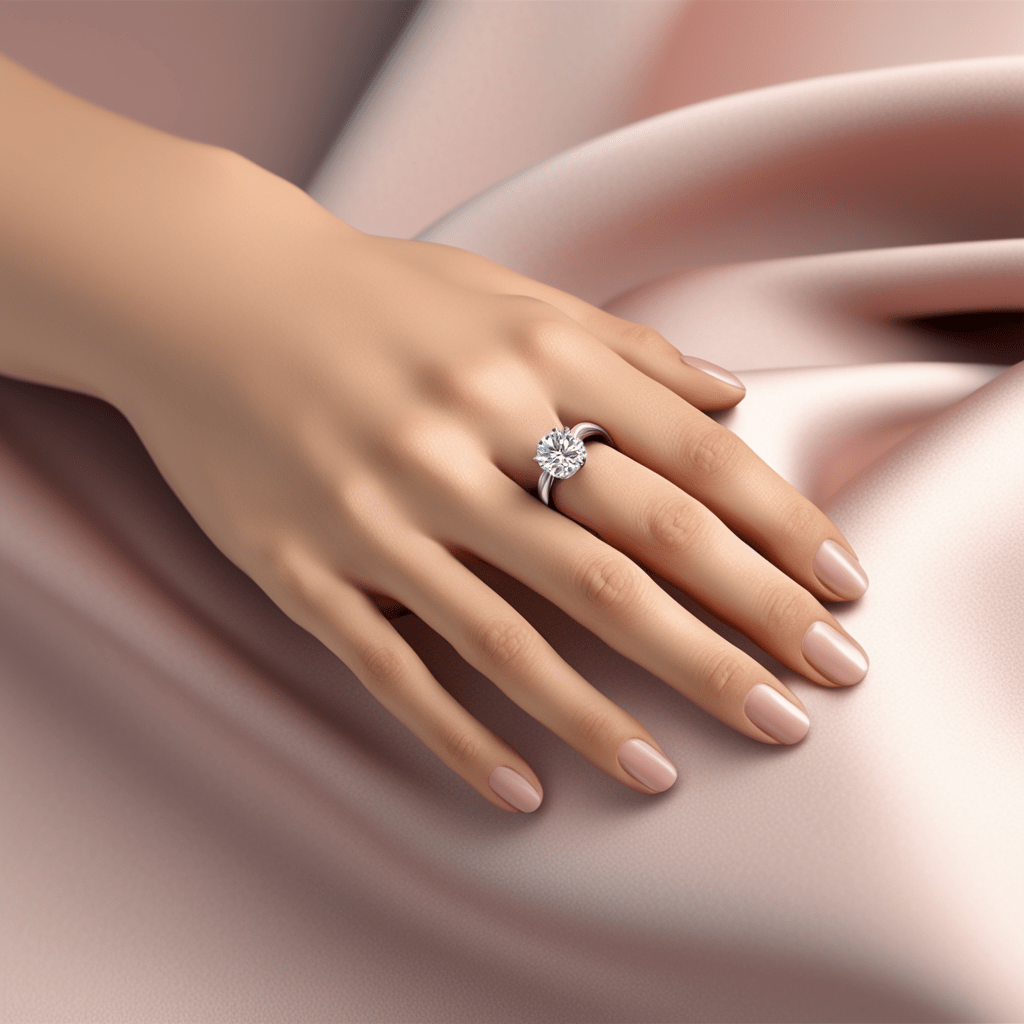} &
\includegraphics[width = 0.2\columnwidth]{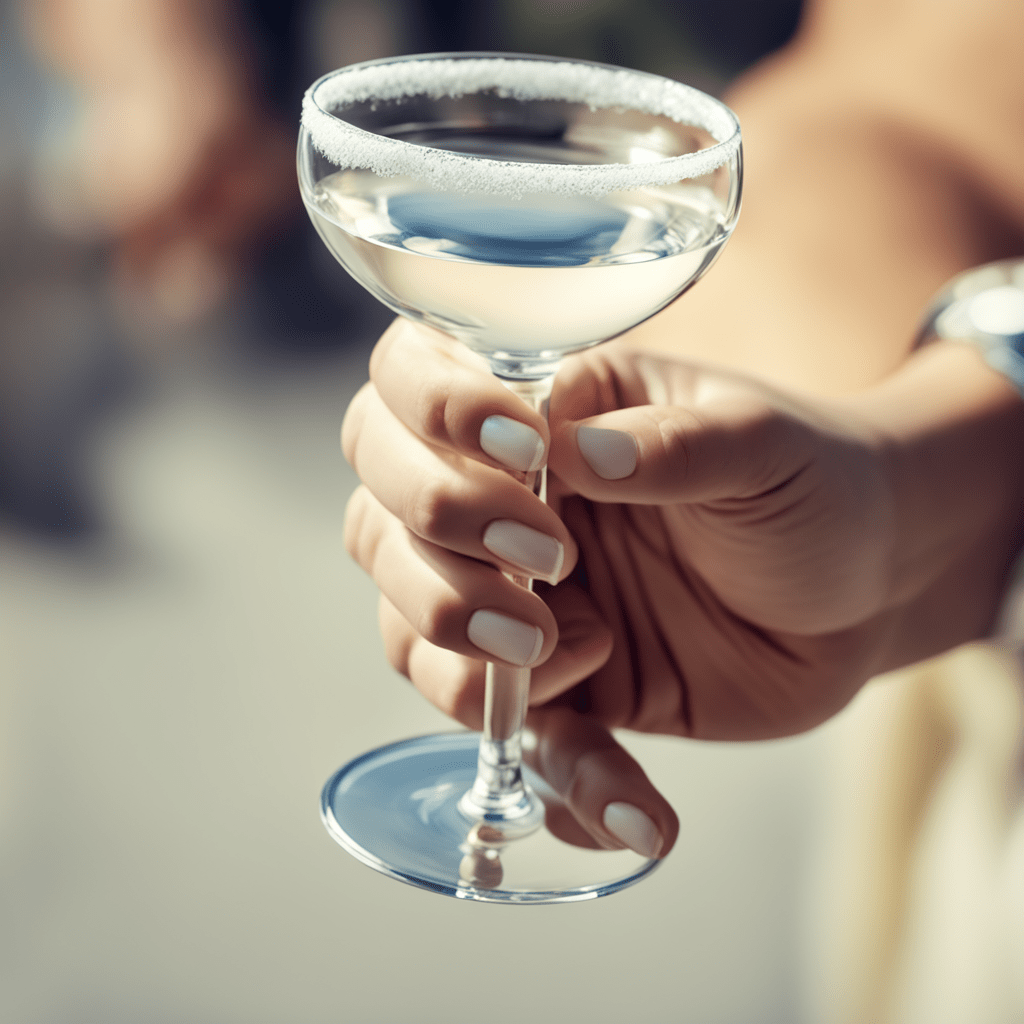} &
\includegraphics[width = 0.2\columnwidth]{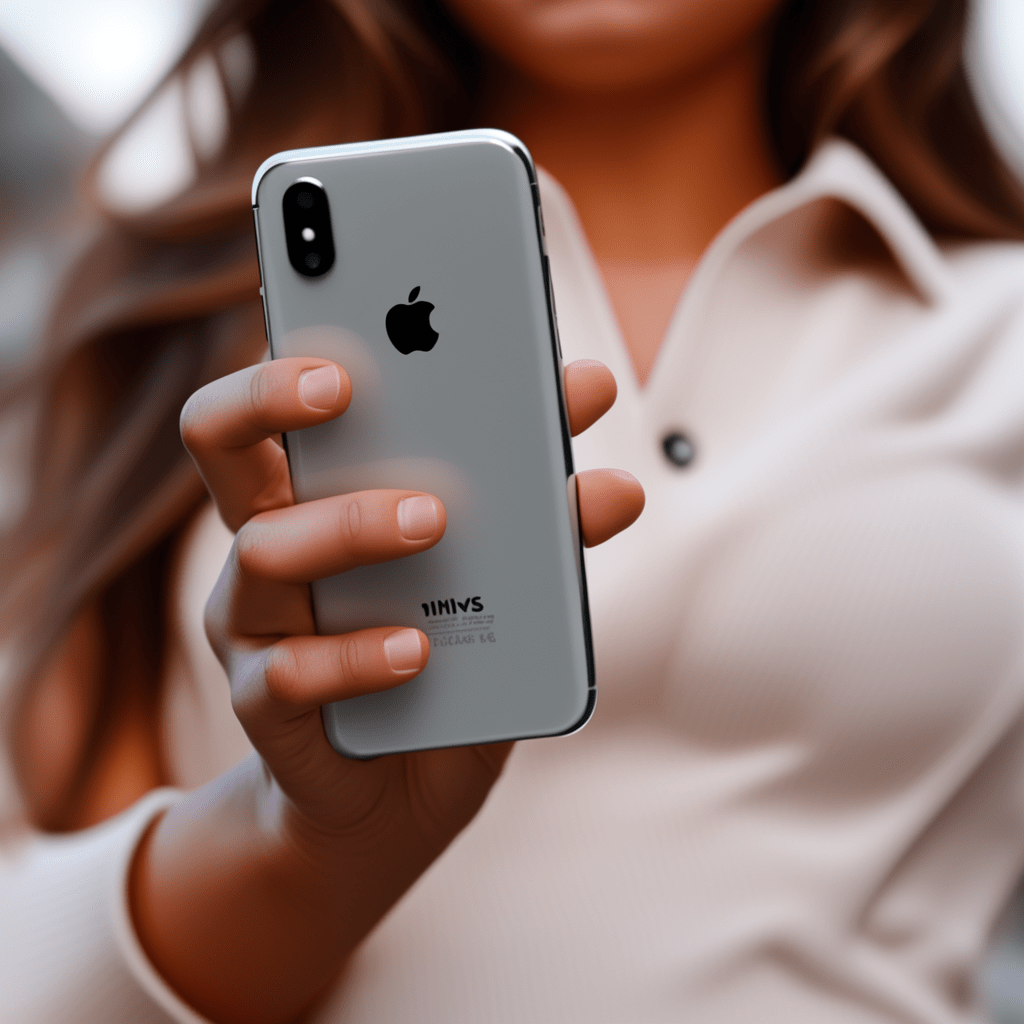} &
\includegraphics[width = 0.2\columnwidth]{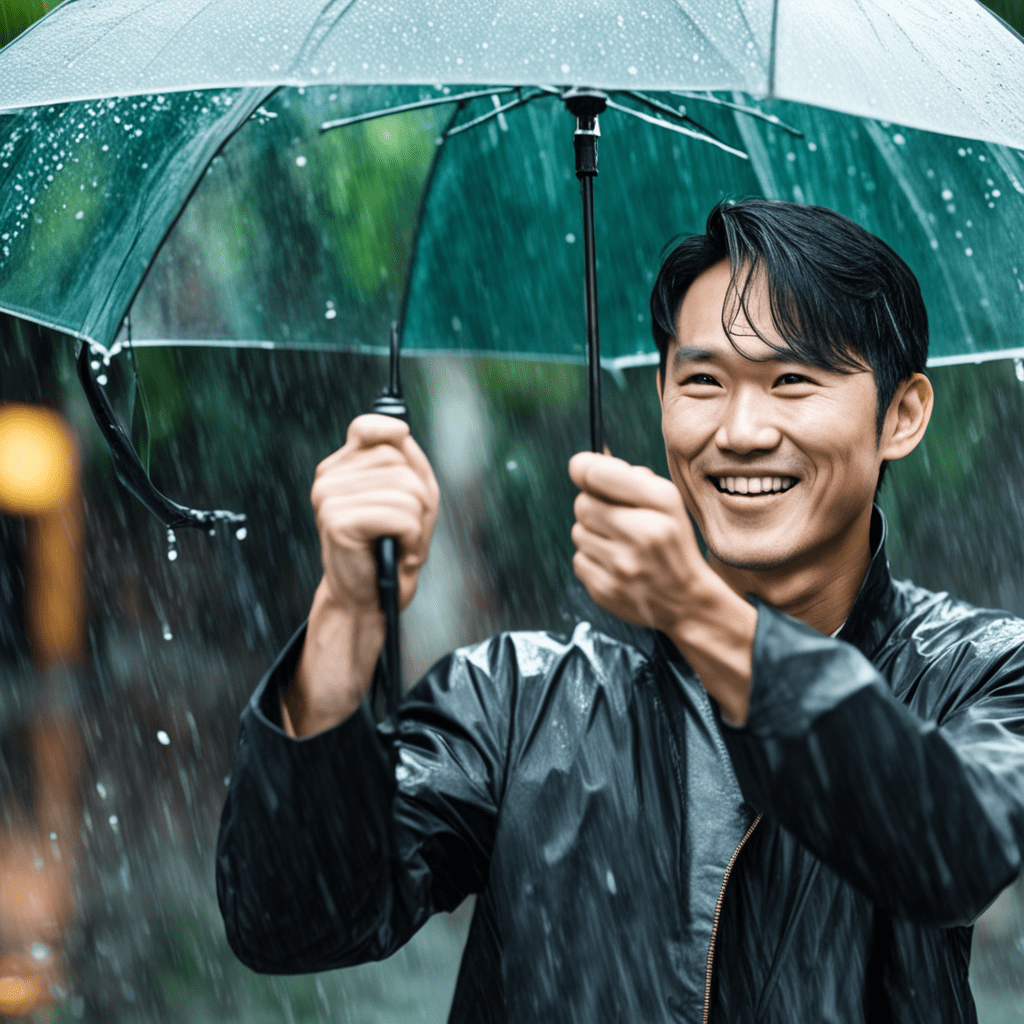}\\
\includegraphics[width = 0.2\columnwidth]{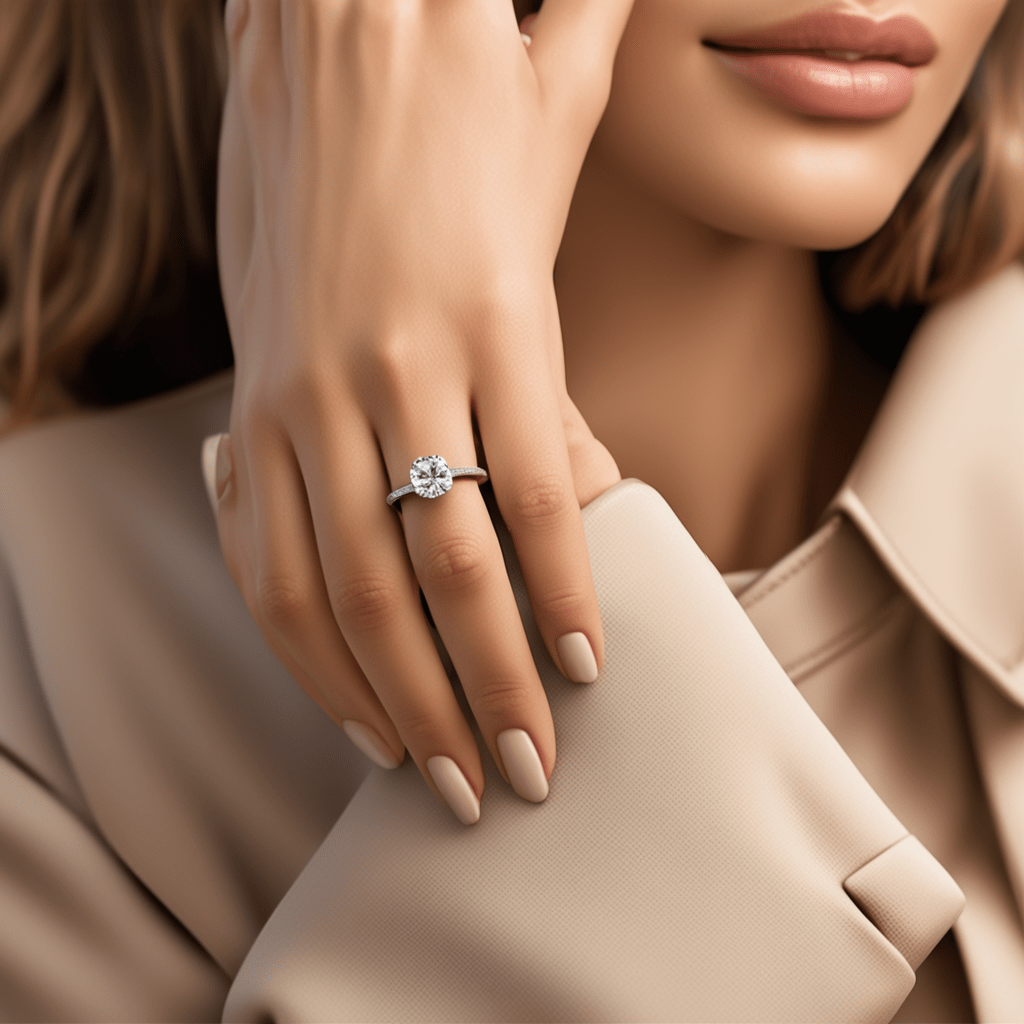} &
\includegraphics[width = 0.2\columnwidth]{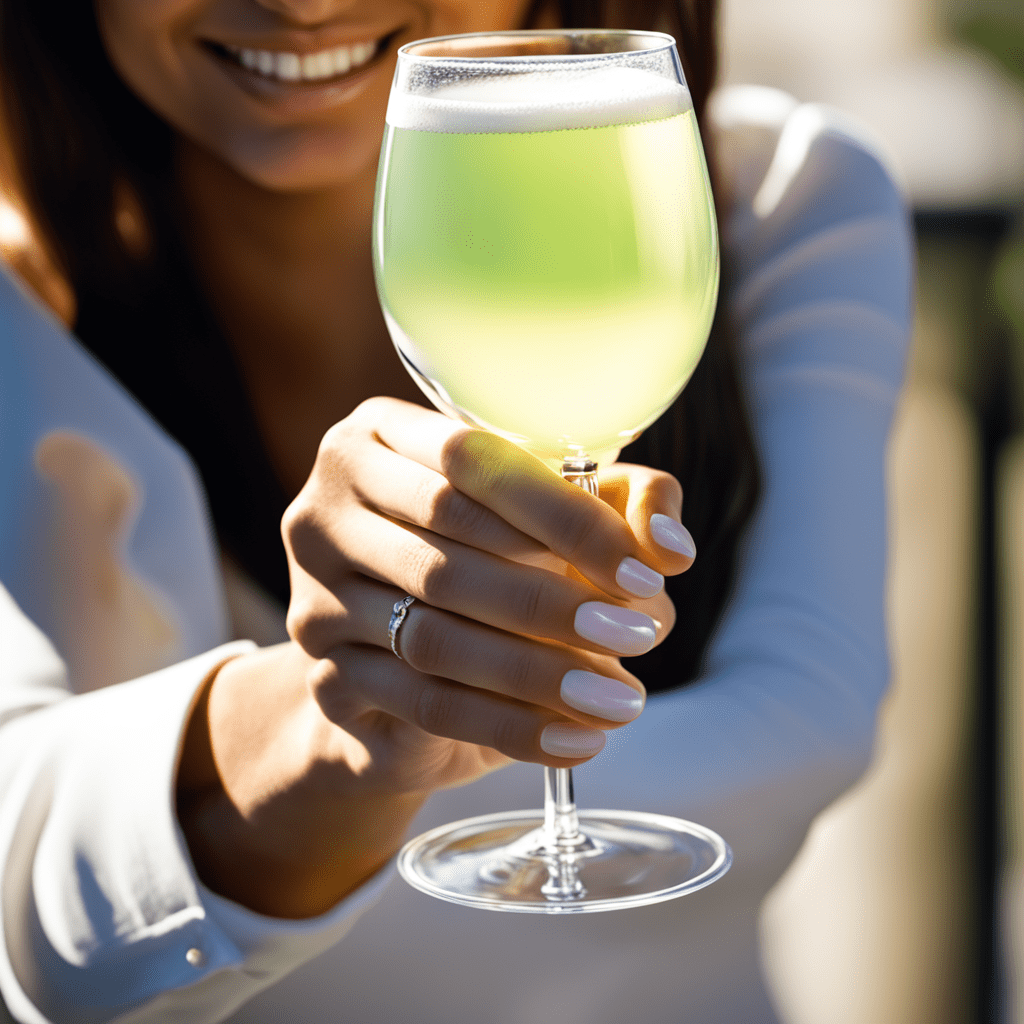} &
\includegraphics[width = 0.2\columnwidth]{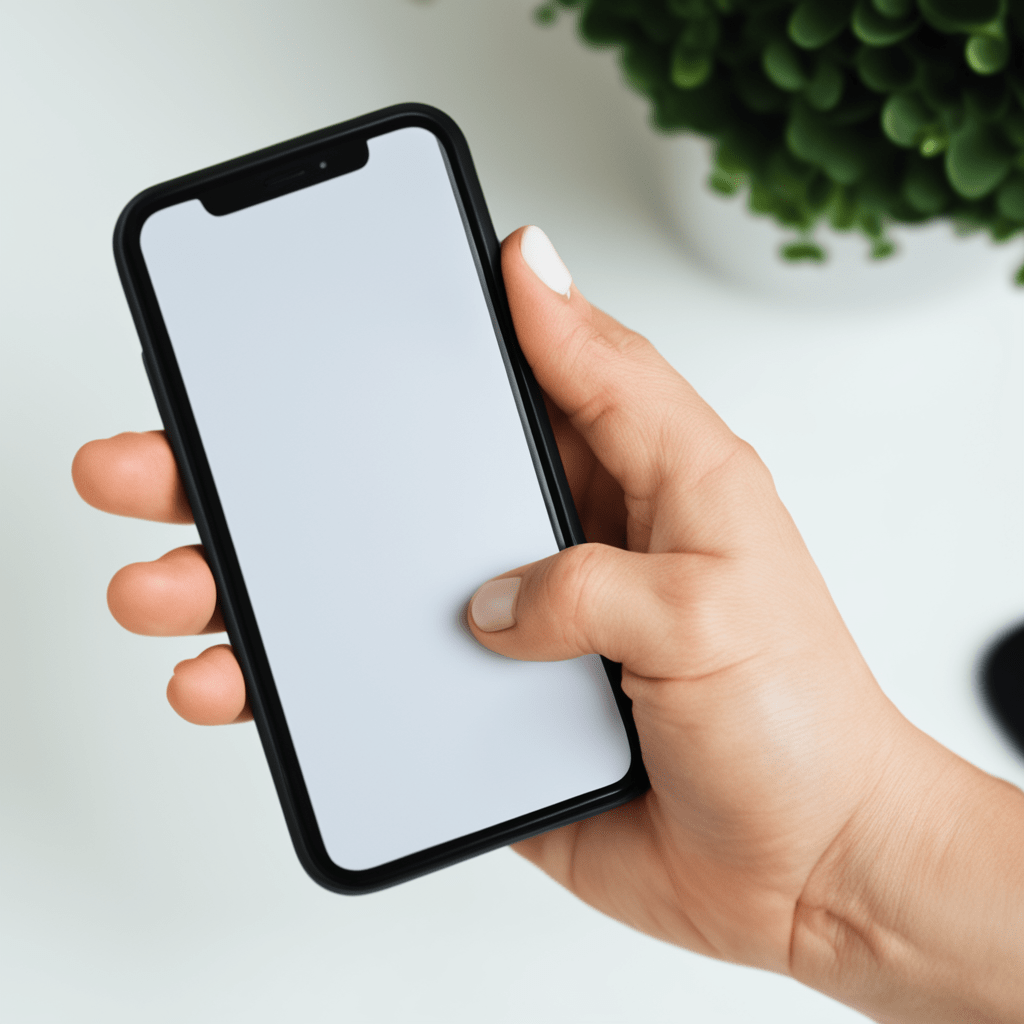} &
\includegraphics[width = 0.2\columnwidth]{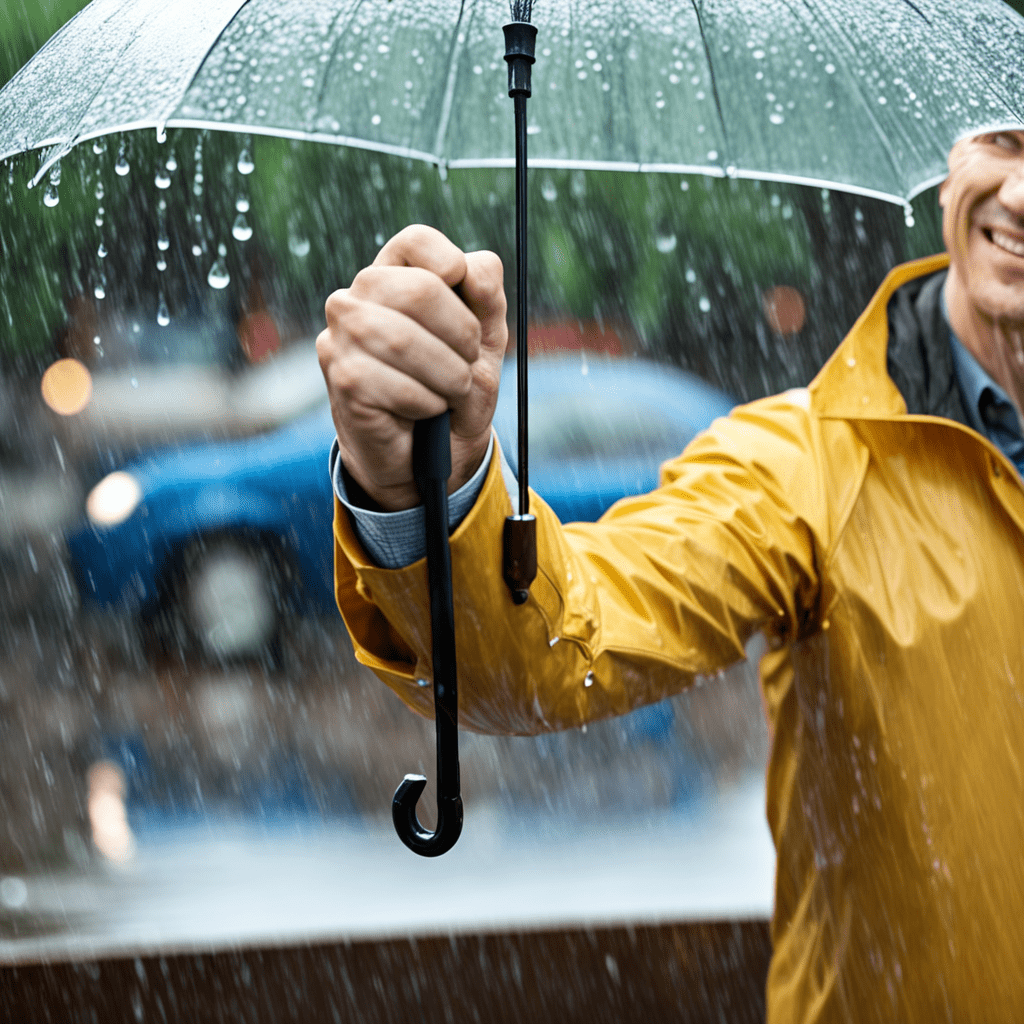}\\
\end{tabular}
\caption{Examples of poor hands being generated by the DreamBooth tuned models.}
\label{poor-examples}
\end{figure}

\section{Training Dataset}
\label{training_dataset}
A few labeled Text-Image pairs are shown in fig \ref{images_with_prompts}. A few cases of poor-hand generation by DreamBooth models are shown in fig \ref{poor-examples}. The breakup of the dataset is given in Table \ref{breakup-table}. Additional examples of the images generated in the \shortname dataset are attached with the supplementary material.

The generic dataset used to fine-tune the SDXL model was the camenduru/hands dataset on HuggingFace.

\section{Failure Modes of framework}
\label{failure-modes}

In this section we highlight a few failure modes of the \framework. The proposer might not exactly follow the prompt exactly or might give disproportionate weight to a few tokens in the initial Prompt $P$, so the generation might not follow the prompt exactly. This issue can be fixed by giving a higher guidance scale during generation or by using something like ReFL \cite{imagereward}
or by creating a better Verifier $\zeta$. The Verifier also might not be able to work well if we have multiple poses and occlusion in the generated images. This can be resolved by creating specific verifiers for proposers so that each verifier is specialized on a generation type and can distinguish between good or bad generations easily. 

The output and generation quality of are framework is also heavily reliant on the effectiveness of the backbone Proposer and Prompter networks used. The output quality of the generated images will vary according to the base diffusion model used. The quality and details of the text pairs corresponding to the images will also differ based on the LM being used.

\section{Meta-Prompt for GPT-4}
\label{meta-prompt-gpt}
The meta-prompt used for generating the prompts and the negative prompts from GPT-4 for the \shortname dataset are given below.

First, for generating the program.
\begin{quote}
    You are given a base sentence for hand-object-interaction. 
    You have to write a detailed program building upon the base sentence. 
    The program contains information about the hands, finger positions and object. 
    Write the program based on the posssible physical positions of the hand, do not write something that cannot be physically executed.
    Following is the syntax of the program. 

    Fully closed meaning finger tips touching palm. 
    Half closed meaning wrapped around the object. 
    Fully open meaning straight. 
    
    [Begin]
    
    Right Hand:
    
        Motion Type: Full Finger Grasp / Full Finger Wrap / Finger Tip Grasp / Support / Lever Grasp / Press / Two Finger Grasp / Three Finger Grasp
        
        Thumb: Fully Open / Half Closed / Fully Closed
        
        Index Finger: Fully Open / Half Closed / Fully Closed
        
        Middle Finger: Fully Open / Half Closed / Fully Closed
        
        Ring Finger: Fully Open / Half Closed / Fully Closed
        
        Little Finger: Fully Open / Half Closed / Fully Closed
        
    Left Hand:
    
        Motion Type: Full Finger Grasp / Full Finger Wrap / Finger Tip Grasp / Support / Lever Grasp / Press / Two Finger Grasp / Three Finger Grasp
        
        Thumb: Fully Open / Half Closed / Fully Closed
        
        Index Finger: Fully Open / Half Closed / Fully Closed
        
        Middle Finger: Fully Open / Half Closed / Fully Closed
        
        Ring Finger: Fully Open / Half Closed / Fully Closed
        
        Little Finger: Fully Open / Half Closed / Fully Closed
        
    Object:
    
        Object Name: Name Of Object
        
        Object Size wrt Hand : Tiny / Small / Size Of Palm / Larger Than Palm
        
        Position wrt Palm: Fully Touching Palm / Not Touching Palm / Partially Touching Palm
        
        Contact with Thumb: Full Thumb/No Thumb/Tip of Thumb/Base of Thumb
        
        Contact with Index Finger: Full Finger/No Finger/Tip of Finger/Base of Finger
        
        Contact with Middle Finger: Full Finger/No Finger/Tip of Finger/Base of Finger
        
        Contact with Ring Finger: Full Finger/No Finger/Tip of Finger/Base of Finger
        
        Contact with Little Finger: Full Finger/No Finger/Tip of Finger/Base of Finger
        
    [End]

    Now write a code for the following base sentence:
    {base}
    Think step by step. First describe the position of hands in text and reason about the positoins and their physical satisfiability then write the code at the end.

\end{quote}

Second, for generating natural language prompt from program.
\begin{quote}
    You are a prompt designer for a text to image system. Text to image systems do not produce ood hands, you have to write a prompt with your main focus on the hands. 
    1. You are given a base code which has the core contents of the prompt. You ar ealso given a base prompt that has additional information about the hand and objects. You need to follow that code and base prompt to produce hand object interaction prompt. 
    2. Write elaborate prompts focusing on the hands. Give the detailed description of finger positions, hand positions, hand gestures, etc. 
    3. Append terms like realistic, 4k, high resolution to the prompt. You also have to write a negative prompt. The negative prompt would include all the things that we do not want in the image. For example, bad hands, extra fingers, broken fingers,
    etc.

    A few examples are given below:
    [Positive Prompt:  A human hand tightly grips a knife, with the index
    and middle fingers around the handle half closed, thumb opposite fully open. The ring and pinky fingers curl inward fully closed. The downward-pointing blade gleams in the light.]
     [Negative Prompt: blurry, disfigured, extra fingers, bad anatomy,too many fingers, cartoon, painting, illus-
    tration,six fingers,(worst quality, low quality, normal quality:2)]
    Now, write a positive and negative prompt for the following input: 
    Input code: {code}
    Input Prompt: {prompt}
    Make sure the prompts that you write are no more
    than 50 words and the object is clearly mentioned in
    the prompt. Give the final prompt in brackets as [pos-
    itive prompt] and [bad prompt].
\end{quote}

\section{Example of DSL Program}
\label{dsl-prog-example}
Sample output of the DSL program generated : 

\begin{quote}
Right\_Hand: \\
    - Motion\_Type: Support \\
    - Thumb: Fully\_Open \\ 
    - Index\_Finger: Fully\_Open \\
    - Middle\_Finger: Fully\_Open \\
    - Ring\_Finger: Fully\_Open \\
    - Little\_Finger: Fully\_Open \\
Left\_Hand: \\
    - Motion\_Type: Full\_Hand\_Grasp \\ 
    - Thumb: Fully\_Closed \\ 
    - Index\_Finger: Fully\_Closed \\ 
    - Middle\_Finger: Fully\_Closed \\
    - Ring\_Finger: Fully\_Closed \\
    - Little\_Finger: Fully\_Closed \\
Object: \\ 
    - Object\_Name: Tea Filled Cup \\
    - Object\_Size\_wrt\_Hand: Size\_Of\_Palm \\
    - Position\_wrt\_Palm: Not\_Touching\_Palm \\
    - Contact\_with\_Thumb: Full\_Thumb \\ 
    - Contact\_with\_Index\_Finger: Full\_Finger \\
    - Contact\_with\_Middle Finger: Full\_Finger \\
    - Contact\_with\_Ring\_Finger: Full\_Finger \\
    - Contact\_with\_Little\_Finger: Full\_Finger \\   
\end{quote}

Based on this prompt, the output generation prompt generated by GPT-4 is as follows :

\begin{quote}
    Right hand is in a supportive position with all fingers fully open. Left hand is in a full grasp motion with all fingers fully curled. The tea filled cup is placed on the right hand. The left hand holds the handle of cup, not touching the palm. The tea gleams in the light with realistic details
\end{quote}

\bibliography{refs}

\end{document}